%% file: main.tex
\documentclass[sn-mathphys-num]{sn-jnl}

\usepackage{graphicx}%
\usepackage{multirow}%
\usepackage{amsmath,amssymb,amsfonts}%
\usepackage{amsthm}%
\usepackage{mathrsfs}%
\usepackage[title]{appendix}%
\usepackage{xcolor}%
\usepackage{textcomp}%
\usepackage{manyfoot}%
\usepackage{booktabs}%
\usepackage{algorithm}%
\usepackage{algorithmicx}%
\usepackage{algpseudocode}%
\usepackage{listings}%

\usepackage{arydshln}
\usepackage{caption}
\usepackage{csquotes}
\usepackage{adjustbox}
\usepackage{lipsum}

\theoremstyle{thmstyleone}%
\theoremstyle{thmstyletwo}%

\theoremstyle{thmstylethree}%

\begin{document}

\title[Meerkat]{Small Language Models Learn Enhanced Reasoning Skills from Medical Textbooks}


\author[1]{\fnm{Hyunjae} \sur{Kim}}\email{hyunjae-kim@korea.ac.kr}

\author[1]{\fnm{Hyeon} \sur{Hwang}}
\email{hyeon-hwang@korea.ac.kr}

\author[1]{\fnm{Jiwoo} \sur{Lee}}
\email{hijiwoo7@korea.ac.kr}

\author[1]{\fnm{Sihyeon} \sur{Park}}
\email{sh10@korea.ac.kr}

\author[1]{\fnm{Dain} \sur{Kim}}
\email{dain-kim@korea.ac.kr}

\author[1]{\fnm{Taewhoo} \sur{Lee}}
\email{taewhoo@korea.ac.kr}

\author[1]{\fnm{Chanwoong} \sur{Yoon}}
\email{cwyoon99@korea.ac.kr}

\author[1]{\fnm{Jiwoong} \sur{Sohn}}
\email{jw\_sohn@korea.ac.kr}




\author[2]{\fnm{Donghee} \sur{Choi}}
\email{donghee.choi@imperial.ac.uk}


\author*[1,3]{\fnm{Jaewoo} \sur{Kang}}\email{kangj@korea.ac.kr}

\affil[1]{
\orgname{Korea University}, 
\orgaddress{\street{145 Anam-ro, Seongbuk-gu}, \city{Seoul}, \postcode{02841}, \country{Republic of Korea}}}


\affil[2]{
\orgname{Imperial College London}, \orgaddress{\street{South Kensington Campus}, \city{London}, \postcode{SW7 2AZ}, \country{United Kingdom}}}

\affil[3]{ 
\orgname{AIGEN Sciences}, \orgaddress{\street{25 Ttukseom-ro 1-gil, Seongdong-gu}, \city{Seoul}, \postcode{04778}, \country{Republic of Korea}}}

\newcommand{\dc}[1]{{\color{blue}\textbf{[Donghee:} #1\textbf{]}}}


\input{sections/abstract}



\maketitle

\input{commands}

\input{sections/introduction}

\input{sections/results}
\input{sections/discussion}
\input{sections/method}


\backmatter

\bigskip


\bibliography{sn-bibliography}

\clearpage

\begin{appendices}

\section{Extended Data}\label{appendix:extended_data}

\input{tables/six_benchmarks}

\input{tables/case_study_longform_1}
\input{tables/case_study_longform_2}
\input{tables/case_study_longform_3}

\input{tables/cot_generation_prompt}
\input{tables/qa_generation_prompt}
\input{tables/chatdoctor_cleaning_prompt}
\input{tables/dialogue_generation_prompt}




\end{appendices}

\end{document}

%% file: sections/abstract.tex
\abstract{
While recent advancements in commercial large language models (LM) have shown promising results in medical tasks, their closed-source nature poses significant privacy and security concerns, hindering their widespread use in the medical field.
Despite efforts to create open-source models, their limited parameters often result in insufficient multi-step reasoning capabilities required for solving complex medical problems.
To address this, we introduce Meerkat, a new family of medical AI systems ranging from 7 to 70 billion parameters.
The models were trained using our new synthetic dataset consisting of high-quality chain-of-thought reasoning paths sourced from 18 medical textbooks, along with diverse instruction-following datasets.
Our systems achieved remarkable accuracy across six medical benchmarks, surpassing the previous best models such as MediTron and BioMistral, and GPT-3.5 by a large margin. 
Notably, Meerkat-7B surpassed the passing threshold of the United States Medical Licensing Examination (USMLE) for the first time for a 7B-parameter model, while Meerkat-70B outperformed GPT-4 by an average of 1.3\%. 
Additionally, Meerkat-70B correctly diagnosed 21 out of 38 complex clinical cases, outperforming humans' 13.8 and closely matching GPT-4's 21.8.
Our systems offered more detailed free-form responses to clinical queries compared to existing small models, approaching the performance level of large commercial models. 
This significantly narrows the performance gap with large LMs, showcasing its effectiveness in addressing complex medical challenges.
}
\keywords{AI for medicine, small language model, chain-of-thought reasoning, synthetic data, USMLE}

%% file: commands.tex
\newcommand{\draftonly}[1]{#1}

\newcommand{\draftcomment}[3]{\draftonly{\textcolor{#2}{{\textbf{[#3 --\textsc{#1}]}}}}}
\newcommand{\todo}[1]{\draftcomment{TODO}{red}{#1}}
\newcommand{\hyunjae}[1]{\draftcomment{hyunjae}{cyan}{#1}}

\newcommand{\ours}{\texttt{Meerkat}}

\interfootnotelinepenalty=10000

%% file: sections/introduction.tex
\section{Introduction}

Recent advancements in large language models (LM) suggest a promising future for the application of artificial intelligence (AI) in the field of healthcare and medicine, serving as efficient and rapid decision-making assistants for professionals~\cite{thirunavukarasu2023large,tian2024opportunities}. 
Several models have exceeded a passing threshold of 60\% on the United States Medical Licensing Examination (USMLE) questions~\cite{kung2023performance,singhal2023large,singhal2023towards,nori2023capabilities,brin2023comparing}, recently reaching a remarkable accuracy rate of 91.1\%~\cite{saab2024capabilities}.
Furthermore, their competency has been showcased in effectively addressing real-world clinical case challenges, including responding to clinical inquiries related to daily practices, engaging in conversational history-taking, and diagnosing complex clinical cases~\cite{zakka2024almanac,tu2024towards,eriksen2023use}.

Despite these achievements, applying large LMs to the medical field is hindered by privacy and security concerns~\cite{li2017exploration,bartoletti2019ai,thirunavukarasu2023large,mesko2023imperative}. 
Proprietary models like OpenAI's GPT-3.5~\cite{openai2022chatgpt} and GPT-4~\cite{achiam2023gpt} can be prone to security vulnerabilities, as users are required to transmit and receive data through web-based APIs\footnote{Application Programming Interface} because of their closed-source nature.
Managing sensitive patient data through services provided by commercial companies becomes a challenge in the absence of well-established regulations.

Recent efforts have been made to develop open-source models that can be deployed on in-house servers~\cite{touvron2023llama,jiang2023mistral,google2024gemma} and to specialize them for the biomedical and clinical domains~\cite{wu2023pmc,chen2023meditron,labrak2024biomistral,xie2024me}.
However, the problem remains that these models lack the necessary multi-step reasoning capabilities to solve complex problems.
In medicine, strong reasoning skills are particularly crucial for analyzing problems systematically, constructing logical paths, and accurately predicting answers.
While commercial large LMs naturally exhibit this ``chain-of-thought'' (CoT) reasoning ability~\cite{wei2022chain}, thanks to their vast amount of parameters often exceeding 100 billion, smaller open-source LMs do not inherently acquire these abilities during pre-training~\cite{wei2022emergent,tay2022unifying,chung2022scaling}.
This necessitates the development of an open-source model that possesses adequate medical reasoning capabilities to tackle complex medical problems.

\input{figures/overview}

In this study, we introduce Meerkat, a new family of open-source medical AI systems with enhanced reasoning skills acquired from textbooks.
Our model is built upon the current state-of-the-art LMs, such as Mistral-7B~\cite{jiang2023mistral}, LLaMA-3-8B, and LLaMA-3-70B~\cite{jiang2023mistral}, and fine-tuned using a diverse set of carefully crafted data. 
Specifically, we used 9.3K USMLE-style questions with corresponding CoT reasoning paths from the MedQA dataset~\cite{jin2021disease}, along with 78K high-quality synthetic CoT data generated from 18 medical textbooks. 
We also utilized instruction-following and chat datasets to address a wide range of use cases in this field.
Meerkat-7B and Meerkat-8B achieved an average accuracy of 62.6\% and 67.3\% across six benchmarks, surpassing GPT-3.5 (54.8\%)~\cite{openai2022chatgpt}, MediTron-7B (51.0\%)~\cite{chen2023meditron}, and BioMistral-7B (55.4\%)~\cite{labrak2024biomistral}.
Notably, Meerkat-7B achieved scores of 77.1 on the MedQA~\cite{jin2021disease}, marking the first instance where a 7B model surpassed the USMLE's passing threshold of 60\% accuracy.
Our largest model, Meerkat-70B, outperformed GPT-4 by an average of 1.3\%, establishing a new state-of-the-art performance benchmark.
In a test of 38 complex clinical cases, Meerkat-70B accurately diagnosed 21 cases, surpassing the human average of 13.8 and nearly matching GPT-4's score of 21.8.
Furthermore, when comparing the free-form responses of Meerkat and baseline models to day-to-day clinical queries~\cite{manes2024k}, our models delivered more detailed and comprehensive responses compared to existing models of sizes 7B and 13B, and GPT-3.5. 
We underscore that our Meerkat models, along with the CoT fine-tuning approach, substantially narrowed the performance gap with commercial LMs, enabling smaller models to tackle challenging reasoning tasks (Figure~\ref{fig:overview}).
Our contributions are summarized as follows:
\begin{itemize}
    \item We introduce Meerkat, a cutting-edge series of medical AI models with high-level multi-step reasoning capabilities. 
    Meerkat-7B is the first 7B model to exceed the USMLE passing threshold, setting the standard as the leading open-source model in its class. Additionally, Meerkat-70B outperformed GPT-4 by 1.3\% across six medical benchmarks. Moreover, in diagnosing complex real-world clinical cases, Meerkat-70B has demonstrated comparable efficacy to GPT-4.
    \item We will publicly release all the artifacts including our model weights\footnote{\url{https://huggingface.co/dmis-lab/meerkat-7b-v1.0}}\footnote{\url{https://huggingface.co/dmis-lab/llama-3-meerkat-8b-v1.0}}\footnote{\url{https://huggingface.co/dmis-lab/llama-3-meerkat-70b-v1.0}} and training data online, which includes the CoT data for the MedQA questions and the new \texttt{MedBooks-CoT-18} dataset that comprises synthetic question-answer pairs with CoT reasoning paths extracted from 18 textbooks.
    \item To the best of our knowledge, Meerkat represents the first instance of training a medical AI system using CoT data synthesized from raw textbooks and showing its effectiveness (see Methods for details).
\end{itemize}

%% file: figures/overview.tex
\begin{figure*}[t]
\centering
\includegraphics[width=\linewidth]{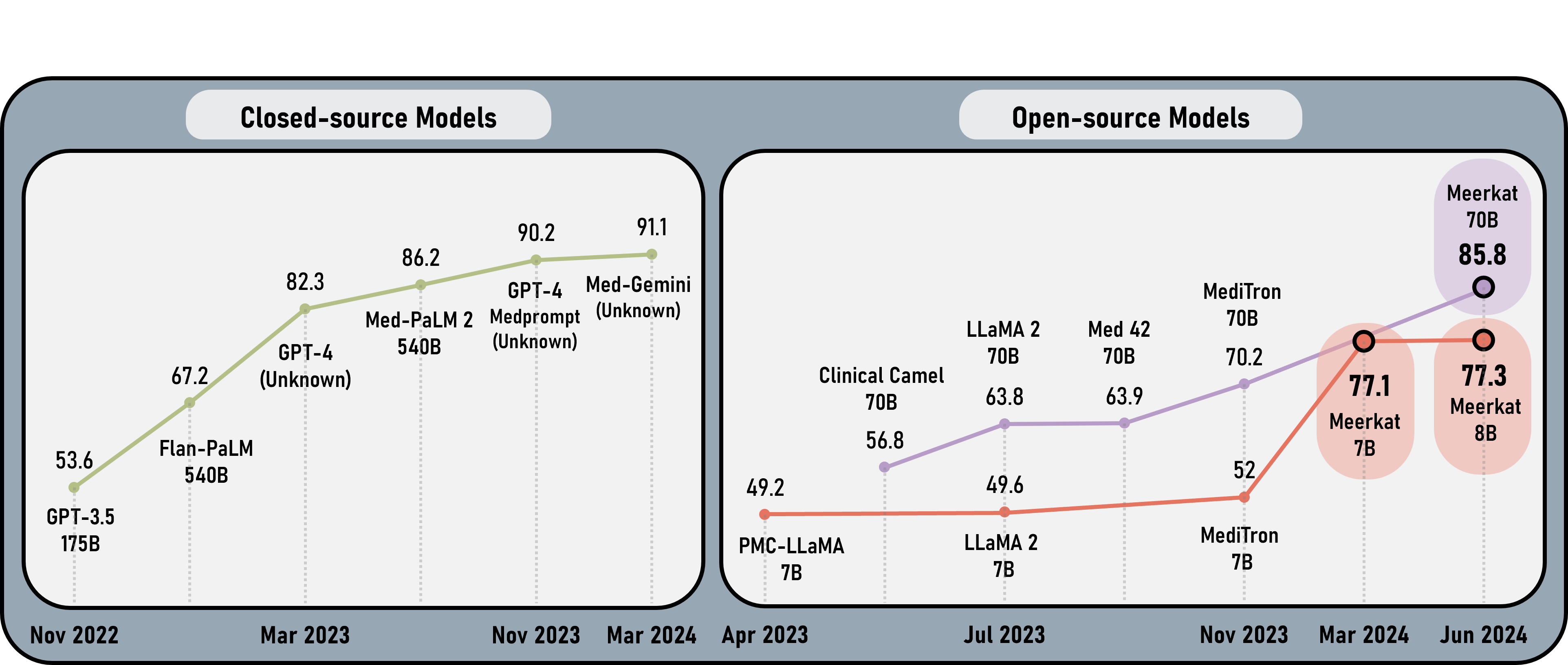}
\vspace{0.5mm}
\caption{
Overview of recent advances in language models (LM) based on their performance on the MedQA benchmark~\cite{jin2021disease}.
Large closed-source models have surpassed the USMLE passing threshold, reaching a state-of-the-art performance with 91.1\% accuracy~\cite{saab2024capabilities}. 
On the other hand, the previous best open-source model, MediTron-70B~\cite{chen2023meditron}, has achieved a score of only 70.2\%, while no 7B-scale models have managed to surpass the USMLE passing threshold (60\%).
Our new open-source model, Meerkat-7B, has achieved an accuracy of 77.1\%, demonstrating notable progress in open-source model development in the medical domain.
Additionally, our new 8B and 70B models have further pushed the state-of-the-art performance for open-source medical AIs.
}
\label{fig:overview}
\end{figure*}

%% file: sections/results.tex
\section{Results}

\subsection{Main Results}
\label{subsec:main_results}

\input{figures/main_results}

\subsubsection{Multiple-Choice QA}
We employed six benchmarks consisting of multiple-choice medical exam questions, which have been commonly used as a fundamental testbed for evaluating LMs in the medical domain~\cite{kung2023performance,singhal2023large,singhal2023towards,nori2023capabilities,brin2023comparing,nori2023can,saab2024capabilities,chen2023meditron,labrak2024biomistral,chen2024benchmarking}.
Figure~\ref{fig:main_results} (a) shows Meerkat-7B significantly outperformed the previous best 7B models, MediTron-7B and BioMistral-7B, by an average of 11.6\% and 7.2\%, respectively. 
Meerkat-8B also surpassed the baseline models with an average improvement of 16.3\% and 11.9\%. 
Remarkably, our models outperformed GPT-3.5 by 7.8\% and 12.5\%, even with fewer parameters.
Our larger 70B model surpassed GPT-4 with an improvement of 1.3\%, setting a new standard for state-of-the-art performance.
Figure~\ref{fig:main_results} (b) illustrates a comparison between our models and GPT-3.5, GPT-4, and the previous best open-source medical model, MediTron-70B, on MedQA.
Meerkat-7B achieved scores of 70.6\% and 77.1\% in single-model and ensemble evaluation settings, respectively, significantly surpassing the 60\% passing threshold and outperforming both GPT-3.5 and the MediTron-70B ensemble model. Meerkat-8B reached 74.0\% and 77.3\% in the same settings. Notably, Meerkat-70B scored 82.6\% and 85.8\%, surpassing GPT-4 by 1.2\% and 4.4\%, respectively.

\subsubsection{NEJM Case Challenges}
Evaluating models solely based on medical exam benchmarks may not sufficiently validate the practical utility of the models, because real clinical scenarios demand more specialized medical knowledge and complex reasoning abilities than typical medical exam questions.
Therefore, we utilized NEJM case challenges,\footnote{\url{https://www.nejm.org/case-challenges}} which consists of challenging and unusual real-world clinical cases.
Models are tasked with predicting the most likely diagnoses for these cases, which include extensive text, laboratory results in tables, and imaging results as captions~\cite{eriksen2023use}.
Figure~\ref{fig:main_results} (c) demonstrates that our models consistently outperformed the previous best instruction-tuned models and humans.\footnote{We also tested medical-specialized models, but excluded them due to lower performance. The human score was determined by NEJM journal readers rather than experts~\cite{eriksen2023use}.}
Notably, our best model, Meerkat-70B, performed comparably to GPT-4.
These results indicate that the reasoning abilities that our models have learned from USMLE-style problems are transferable to challenging real-world clinical problems.

\input{tables/kqa_results}

\subsubsection{Real-world Clinical Questions}

While exam questions are valuable for assessing the medical knowledge and reasoning abilities of language models, they may not fully reflect the complexities and breadth of real-world clinical queries.
Therefore, we compared Meerkat models with instruction-tuned LMs using the K-QA dataset~\cite{manes2024k}, which includes real-world clinical questions commonly encountered in clinicians' daily practices, along with corresponding long-form answers annotated by healthcare experts.
Table~\ref{tab:kqa_results} demonstrates that the completeness score of our models significantly surpassed that of all other open-source LMs and was comparable to or better than GPT-3.5, indicating that our models provided detailed responses that encompass necessary parts for a given query.
However, our models exhibit inferior factuality performance compared to GPT-3.5 and GPT-4, implying that the issue of hallucination in small models remains a persistent challenge.
Tables~\ref{tab:case_1},~\ref{tab:case_2},~and~\ref{tab:case_3} present a case study comparing the responses of human experts, GPT-4, ChatDoctor-7B, and Meerkat-7B.

\subsection{Ablation Study}
\label{subsec:ablation}

\input{figures/ablation}

\subsubsection{Effect of CoT Fine-tuning}

To assess the effectiveness of CoT fine-tuning, we compared the performance of models trained solely on question-answer pairs from the MedQA training set (referred to as ``MedQA'') with those trained on both MedQA data and CoT reasoning (referred to as ``MedQA-CoT'').
Figure~\ref{fig:ablation} demonstrates that CoT fine-tuning dramatically improved MedQA performance by an average accuracy of 7.5\% across the six models.

\subsubsection{Effect of Textbook Augmentation}

Figure~\ref{fig:ablation} shows that augmenting the training data with additional QA pairs and CoT reasoning paths obtained from textbooks (referred to as ``MedQA-CoT \& MedBooks-CoT-18'') led to a further improvement in performance, with an average accuracy increase of 5.1\% across the six models compared to those trained using MedQA-CoT alone.

\subsubsection{Language Model Selection}

We assessed six open-source LMs with 7B or 8B parameters released between July 2023 and April 2024 using the MedQA dataset. 
As depicted in Figure~\ref{fig:ablation}, general-purpose models like Mistral-7B, Gemma-7B~\cite{google2024gemma}, and LLaMA-3-8B outperformed biomedical-specific models such as MediTron-7B and BioMistral-7B.
Although the details of their pretraining corpus remain unknown, we hypothesize that these models achieved high MedQA performance due to extensive training on diverse corpora, likely including a large amount of biomedical literature.
While Mistral and Gemma exhibited similar performance, we selected Mistral-7B as our backbone model because of its faster inference speed~\cite{jiang2023mistral}.
Following the release of LLaMA-3 in April 2024, we proceeded with training a new set of models, resulting in Meerkat-8B and Meerkat-70B.

\subsection{Assessment of Model Explanations}
\label{sec:eval_reasoning}

\input{figures/reasoning_eval}

We evaluated the quality of explanations for USMLE-style questions in the Medbullets-4 dataset from Mistral-7B-Instruct, GPT-4, and Meerkat-7B by comparing them to human explanations.\footnote{These human explanations are sourced from the Medbullets dataset~\cite{chen2024benchmarking} and offer in-depth analyses of the questions and both correct and incorrect options.}
For evaluation, we employed two standard scoring metrics, ROUGE-L~\cite{lin2004rouge} and BERTScore~\cite{zhang2019bertscore}.
Additionally, we calculated the GPT-4 score, assigning a score from 1 to 5 to model explanations based on completeness, accuracy, clarity, relevance, and logical coherence, with human explanations assumed to receive a score of 5.
Figure~\ref{fig:reasoning_eval} illustrates that Meerkat-7B exhibits the highest performance compared to the other instruct-tuned models based on ROUGE-L and BERTScore metrics, indicating that Meerkat-7B provided explanations that are lexically and semantically similar to human explanations. 
Based on the GPT-4 score, GPT-4 achieved the highest score, while Meerkat-7B fell short compared to GPT-4 but outperformed Mistral-7B-Instruct.
Furthermore, the score of explanations provided by Meerkat-7B for correctly answered questions was over 1.4 points higher than the score for explanations of incorrect answers, indicating that the quality of reasoning paths highly influences the accuracy of answers~\cite{zhang2023language}.

%% file: figures/main_results.tex
\begin{figure*}[t]
\centering
\includegraphics[width=\linewidth]{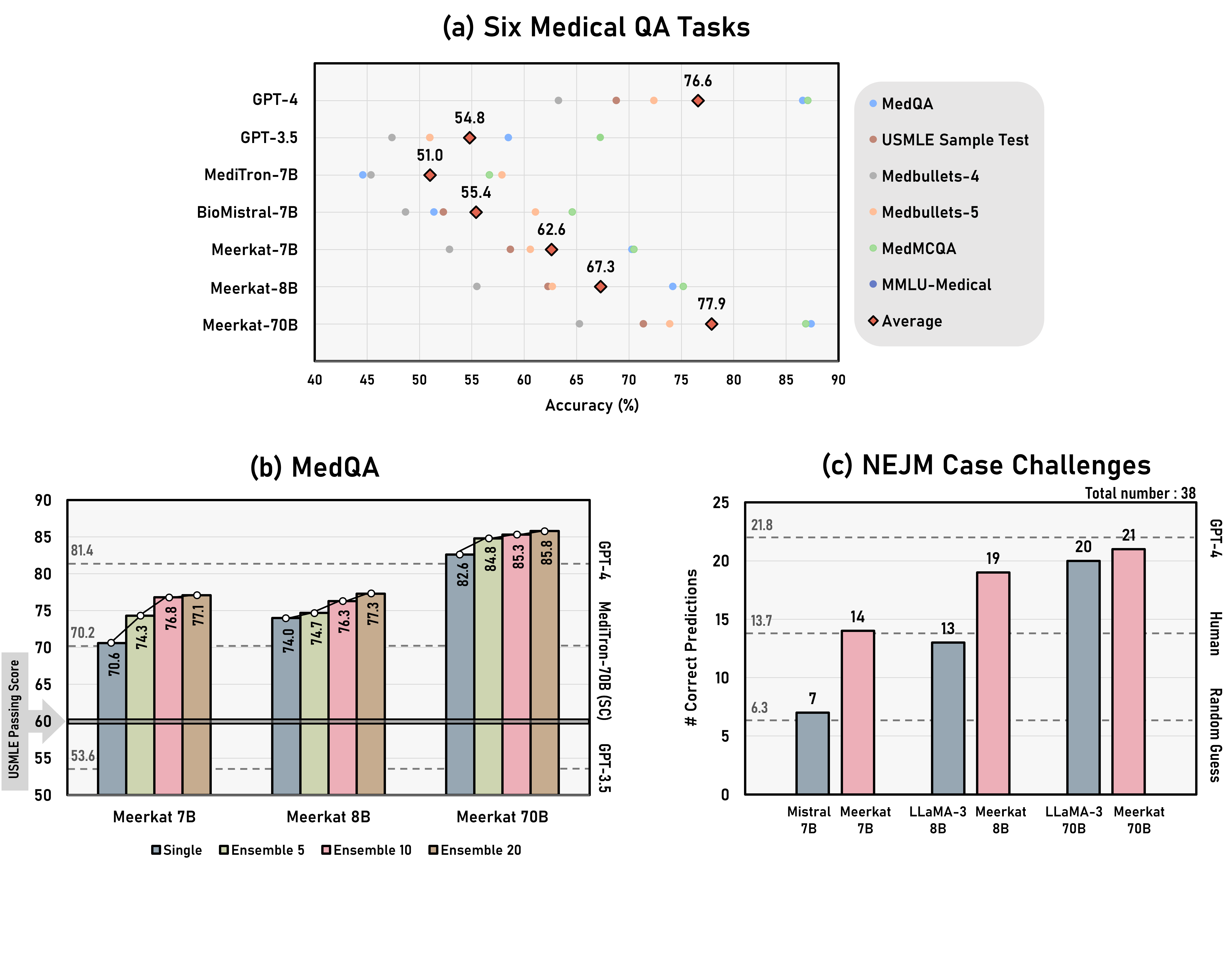} 
\vspace{-12mm}
\caption{
Performance of models on six multiple-choice QA benchmark datasets and NEJM case challenges.
Our Meerkat models generally performed better than existing 7B models and GPT-3.5 across the six datasets and outperformed MediTron-70B on MedQA.
Additionally, our 70B model exceeded GPT-4 in performance.
The scores of GPT-3.5, GPT-4 and MediTron-70B are obtained from the papers of \citet{nori2023capabilities}, \citet{toma2023clinical}, \citet{chen2023meditron}, and \citet{chen2024benchmarking}.
In subfigure (b), MediTron-70B (SC) denotes that the self-consistency CoT prompting method was employed during the model's evaluation~\cite{wang2023self}, whereas Ensemble 5-20 refers to the number of runs of choice shuffling ensemble~\cite{nori2023capabilities}.
Detailed scores for the six benchmarks shown in subfigure (a) are provided in Table~\ref{tab:six_benchmarks}.
}
\label{fig:main_results}
\end{figure*}

%% file: tables/kqa_results.tex
\begin{table}[t]
\captionsetup{width=\textwidth}
\caption{
The performance of models in providing long-form responses for everyday clinical questions in the K-QA dataset~\cite{manes2024k}.
The completeness score (\%) measures how well the response includes essential content from the gold standard response for a given query.
The factuality score (\%) evaluates whether the model's response contradicts the gold standard response for a given query
(see Section~\ref{subsec:experimental_settings} for details of the metrics).
We chose not to assign a factuality score to Med-Alpaca, despite achieving a high factuality score, because it often generates irrelevant responses to the given query and these low-quality responses are not considered in the factuality score.
}\label{tab:kqa_results}
\begin{tabular}{lrcc}
\toprule
\textbf{Model} & \textbf{Size} & \textbf{Completeness} & \textbf{Factuality} \\
\midrule
GPT-4 & Unknown & 81.0 & 92.5  \\
GPT-3.5 & 175B & 71.4 & 92.0  \\
\midrule
ChatDoctor~\cite{li2023chatdoctor} & 7B & 63.0 & 89.1  \\
Mistral-Instruct~\cite{jiang2023mistral} & 7B & 62.4 & 88.1  \\
Med-Alpaca~\cite{han2023medalpaca} & 13B & 6.8 & -  \\
PMC-LLaMA~\cite{wu2023pmc} & 13B & 49.8 & 90.0 \\
\midrule
Meerkat (\textbf{Ours}) & 7B & 70.3 & 89.6  \\
 & 8B & 72.2 & 90.0  \\
 & 70B & 75.4 & 89.6  \\
\bottomrule
\end{tabular}
\end{table}

%% file: figures/ablation.tex
\begin{figure*}[t]
\centering
\includegraphics[width=0.99\linewidth]{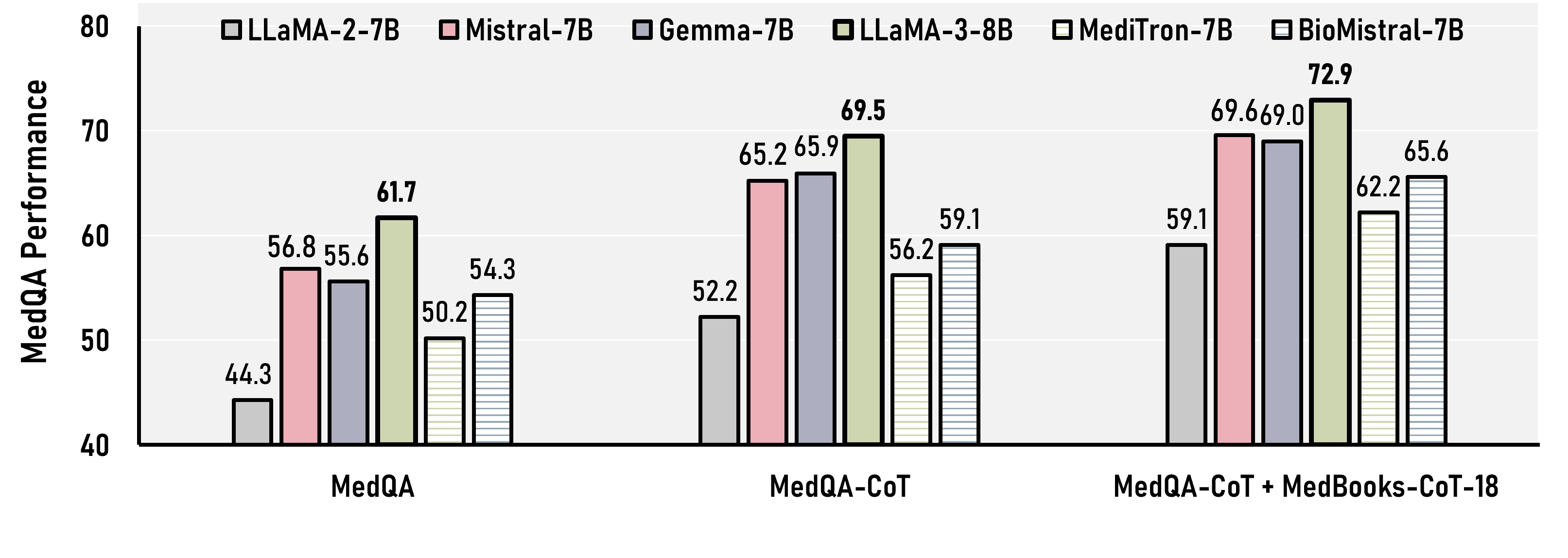} 
\caption{
Performance comparison of six language models trained with three different datasets on the MedQA benchmark.
Mistral-7B, Gemma-7B~\cite{google2024gemma}, and LLaMA-3-8B performed better than MediTron-7B and BioMistral-7B, despite not being specialized models for biomedicine.
``MedQA'': training the model only using question-answer pairs in the MedQA training set.
``MedQA-CoT'': training the model using MedQA question-answer pairs and corresponding CoT reasoning data.
``MedQA-CoT + MedBooks-CoT-18'': training the model using the MedQA-CoT data and additional CoT data generated from textbooks.
}
\label{fig:ablation}
\end{figure*}

%% file: figures/reasoning_eval.tex
\begin{figure*}[t]
\centering
\includegraphics[width=\linewidth]{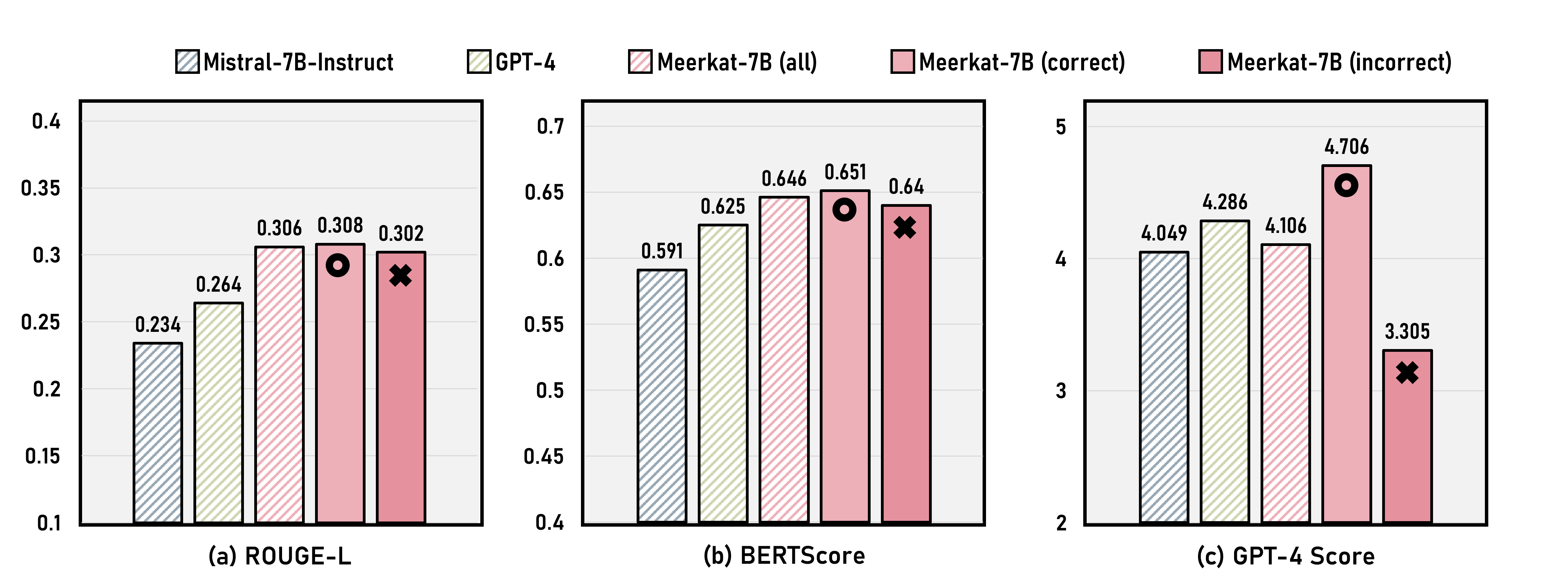}
\vspace{0.5mm}
\caption{
Evaluation of model explanations for USMLE-style questions.
the scores were measured by comparison with human explanations.
Meerkat-7B performed the best according to ROUGE-L~\cite{lin2004rouge} and BERTScore~\cite{zhang2019bertscore}, and it ranked second in terms of the GPT-4 score.
``O'' denotes explanations for questions that Meerkat-7B answered correctly, while ``X'' indicates explanations for questions the model answered incorrectly.
}
\label{fig:reasoning_eval}
\end{figure*}

%% file: sections/discussion.tex
\section{Discussion}

In this study, we introduced a new series of medical AI systems called Meerkat. 
The models are small LMs ranging 7 to 70 billion parameters and are openly available as open-source software, allowing it to run on-premise with relatively low-spec GPUs (e.g., a single 24G NVIDIA GeForce RTX 3090 GPU for the 7B and 8B models) in research labs and hospitals.  
One of our key contributions is the creation of the MedBooks-CoT-18 dataset, where we utilized GPT-4 to generate USMLE-style questions from medical textbooks and collected multi-step reasoning paths required to answer these questions.
As a result of fine-tuning our models with this high-quality CoT data, the models largely outperformed the baseline models across six benchmarks and NEJM case challenges.

As observed in Table~\ref{tab:kqa_results}, our models provided more detailed responses without sacrificing factuality compared to the baseline small models. 
However, when compared to larger LMs such as GPT-3.5 and GPT-4, they exhibited a noticeable decrease in factuality score, suggesting the need for further development towards more reliable AI systems in future research.
Furthermore, given that our models were not fine-tuned using preference alignment techniques like reinforcement learning from human feedback (RLHF)~\cite{ouyang2022training}, there is a possibility it could offer unsupported, unsafe, or biased responses. 
Hence, it is crucial to exercise caution and obtain expert validation before deploying the models in real-world scenarios to guarantee their reliability.
In this study, we assessed models for diagnosing clinical cases, focusing specifically on constrained clinical scenarios.
In future research, it would be intriguing to advance these models further and thoroughly investigate whether they could effectively assist experts across various clinical settings.

%% file: sections/method.tex
\section{Methods}

\input{tables/train_sets}

Meerkat models are based on state-of-the-art open-source models such as Mistral-7B and LLaMA-3 and are specifically instruct-tuned for the medical domain. 
The key highlight of the model training lies in constructing a high-quality instruction-tuning dataset, which is elaborated in Section~\ref{subsec:training_details}.
Details of experimental settings including descriptions of benchmark datasets can be found in Section~\ref{subsec:experimental_settings}.

\subsection{Training}
\label{subsec:training_details}

The training data preparation involves three primary steps: generating high-quality CoT data from a question-answering (QA) dataset, augmenting this CoT data using medical textbooks, and reusing/repurposing pre-existing instruction-following datasets covering various medical use cases with suitable modifications.
Table~\ref{tab:train_sets} lists the datasets used for training the Meerkat models.
Figure~\ref{fig:framework} depicts the overall process for generating and augmenting CoT data.

\input{figures/framework}

\subsubsection{Chain-of-thought Reasoning Data Generation}

Chain-of-thought (CoT) reasoning~\cite{wei2022chain} refers to the process of providing step-by-step explanations to arrive at a conclusion for complex problems.
Several studies have demonstrated that generating CoT using large LMs and then leveraging it to fine-tune smaller LMs is promising, but it remains understudied in the medical domain.
We prompted GPT-4 to solve USMLE-style questions from the MedQA training set using step-by-step reasoning (see Table~\ref{tab:prompt_cot_gen} for the input prompt).
These questions are specifically designed to evaluate the multi-step medical reasoning skills of humans, making them suitable sources for obtaining CoT reasoning data.
We kept questions if the responses followed the specified output format and the answer predictions were correct; otherwise, we filtered them out, resulting in 9.3K out of 10K questions remaining.
We did not utilize explanations from human experts unlike the study of \citet{singhal2023towards} because GPT-4 offers more detailed explanations for the same questions compared to humans.
Figure~\ref{fig:cot_comparison} provides
examples of explanations from humans and GPT-4.

\input{figures/cot_comparison}

\subsubsection{Data Augmentation With Medical Textbooks}

Collecting CoT paths solely from QA pairs within a single MedQA dataset does not provide a wide enough variety of training examples to maximize the reasoning abilities of small LMs.
To overcome this obstacle, we constructed MedBooks-CoT-18, a dataset containing an additional 78K question-answer pairs along with corresponding CoT paths, which are automatically generated from 18 English medical textbooks, spanning various medical disciplines.\footnote{The textbooks cover anatomy, biochemistry, cell biology, first aid, gynecology, histology, immunology, internal medicine, neurology, obstetrics, pathology, pediatrics, pharmacology, physiology, psychiatry, and surgery. They are provided by the study of \citet{jin2021disease} and released under license for research use. Please refer to the MedQA official GitHub repository for more information (\url{github.com/jind11/MedQA?tab=readme-ov-file}).}
Initially, we segmented textbooks into chunks with an average character-level length of up to 4K, 8K, and 12K, while allowing for overlap between chunks. 
Each text chunk was fed into GPT-4 with three USMLE-style questions sampled from MedQA for reference.
GPT-4 was then prompted to generate the correct answer, plausible options, and questions for a case representation based on the provided textbook chunk, which included the patient's demographic information and symptoms (see Table~\ref{tab:prompt_qa_gen} for prompt details).
By instructing GPT-4 to generate questions based on textbooks, as depicted in Figure~\ref{fig:generated_question}, we can reduce potential hallucinations during question generation and ensure that the questions reflect essential medical knowledge covered in the textbooks, relevant to both medical exams and real-world clinical practice.
Additionally, we instructed the model to refrain from generating questions if the provided text chunk contained significant noise or lacked adequate information to create QA examples.
After USMLE-style questions were newly created from textbooks, we followed a similar procedure to generate CoT reasoning paths using GPT-4 as we did with the MedQA dataset. 
We filtered out examples where GPT-4's predicted answers were different from the original answers provided by GPT-4, similar to the concept of round-trip consistency~\cite{alberti-etal-2019-synthetic}.

\input{figures/generated_question}

\subsubsection{Utilization of Existing Instruction-following Datasets}

In addition to the CoT datasets that we constructed, we incorporated existing instruction-following datasets into the model training to enhance the versatility of our model for various medical-domain applications (see Table~\ref{tab:train_sets} for the summary).
We refined or repurposed several datasets to better suit the model training and align with the target applications.
Below are detailed descriptions of each dataset:
\begin{itemize}
    \item 
    MedMCQA~\cite{pal2022medmcqa}: This large dataset comprises exam questions from the two Indian affiliations, AIIMS (All India Institute of Medical Sciences) and NEET PG (National Eligibility cum Entrance Test for Post Graduate courses).
    We leveraged this dataset because it spans a broad spectrum of medical knowledge across 21 subjects, which could complement the medical knowledge of small LMs.
    Although the dataset also includes human explanations for the questions, we did not utilize them because they were too brief and not sufficiently detailed.
    \item 
    LiveQA~\cite{abacha2017overview}: This dataset contains healthcare-related questions received by the U.S. National Library of Medicine (NLM), accompanied by free-form responses from experts.
    The questions span various topics including diseases, drugs, and more, making it ideal for training our model on real-world queries.
    \item MedicationQA~\cite{abacha2019bridging}:
    This dataset comprises consumer questions, particularly focusing on inquiries related to drugs, along with expert responses.
    Since these types of questions constitute a significant portion of healthcare inquiries, they serve as valuable resources for developing practical medical AI models.
    \item
    ChatDoctor-cleaned (ours):
    The data is derived from ChatDoctor~\cite{li2023chatdoctor}, a collection of real patient inquiries and doctor responses obtained from an online medical consultation platform.
    While ChatDoctor provides rich and useful data examples, it also contains noise inherent to online Q\&A platforms, such as greetings or closing remarks by the doctors (e.g., they often begin the response with ``Welcome to Chat Doctor'' or end the response with ``Best wishes, Chat Doctor.''). 
    To address this, we manually created three noisy inputs and corresponding corrected outputs, utilizing them as in-context examples. 
    We then employed GPT-3.5 to remove noise from 112K original responses, resulting in our ChatDoctor-cleaned dataset.
    See Table~\ref{tab:prompt_chatdoctor_cleaning} for the input prompt.
    \item 
    MedQA-dialogue (ours):
    While engaging in multi-turn dialogue with users is a crucial requirement for medical AI, there's a lack of suitable training datasets for this purpose. 
    To fill this gap, we instructed GPT-3.5 to generate conversations by role-playing as both patients and doctors based on MedQA questions and corresponding CoT reasoning.
    In the dialogue, the patient should minimally communicate their symptoms and medical history, while the doctor should guide the conversation, asking follow-up questions to gather a thorough medical history and records.
    We generated 4.8K conversations corresponding to the intended output format from the entire 9.3K MedQA examples, creating the MedQA-dialogue dataset. Details of the input prompt is provided in Table~\ref{tab:prompt_dialogue_gen}.
    \item AlpaCare~\cite{zhang2023alpacare}: This dataset is a collection of synthetic 52K medical-domain instructions generated by GPT-3.5 and GPT-4, similar to the self-instruct approach~\cite{wang2022self,alpaca}.
    We included this dataset to improve the model's generalizability to various user queries and use cases.
\end{itemize}

\subsubsection{Training Details}
\label{sec:training_details}
We initialized our models with the Mistral-7B-v0.1,\footnote{\url{huggingface.co/mistralai/Mistral-7B-v0.1}} Meta-Llama-3-8B-Instruct,\footnote{\url{huggingface.co/meta-llama/Meta-Llama-3-8B-Instruct}} and Meta-Llama-3-70B-Instruct\footnote{\url{huggingface.co/meta-llama/Meta-Llama-3-70B-Instruct}} weights, respectively.
We fine-tuned the models on a combined dataset comprising the nine training datasets listed in Table~\ref{tab:train_sets}, using a standard next-token prediction objective. 
The 7B model was trained based on the FastChat GitHub repository\footnote{\url{github.com/lm-sys/FastChat}} for three epochs with a batch size of 128, a learning rate of 2e-6, a warm-up ratio of 0.04, and a maximum length of 2,048 tokens, using eight 80G A100 GPUs, which took approximately 1.5 days to complete.
The 8B and 70B models were trained on Google TPUs with batch sizes of 128 and 64, and learning rates of 7e-6 and 5e-6, respectively.
We employed FlashAttention~\cite{dao2022flashattention} and the fully sharded data parallel approach (FSDP) for efficiency.

\subsection{Experimental Settings}
\label{subsec:experimental_settings}

\subsubsection{Benchmark Datasets}
\input{tables/test_sets}

We provide an overview of the benchmark datasets in Table~\ref{tab:test_sets} with detailed descriptions below.

\begin{itemize}
    \item 
    MedQA~\cite{jin2021disease}: MedQA is one of the most widely used benchmarks in the medical domain. 
    The dataset consists of USMLE-style questions curated by medical examination experts from various medical question banks (an example of USMLE-style questions can be found in Figure~\ref{fig:cot_comparison}). 
    These questions are structured in a multiple-choice format, with four options provided for each question.
    \item
    USMLE sample test:\footnote{\url{usmle.org/prepare-your-exam}} This resource is an official study material for students preparing for the USMLE, closely mirroring the style and difficulty level of the actual tests.
    Each question is accompanied by a varying number of options, ranging from four to nine.
    We utilized the preprocessed version of this data as provided by \citet{toma2023clinical}.\footnote{\url{github.com/bowang-lab/clinical-camel}}
    \item
    Medbullets~\cite{chen2024benchmarking}: 
    This dataset comprises USMLE-style questions sourced from tweets posted since April 2022. Compared to questions in MedQA or the USMLE sample test, these questions are less likely to have been encountered during pre-training, making them more challenging to solve.
    We utilized both Medbullets-4, which provides four options, and Medbullets-5, which offers five options.
    \item
    MedMCQA~\cite{pal2022medmcqa}: This benchmark corresponds to the test split of the MedMCQA dataset, which consists of medical exam questions with four options.
    Note that we used the training split for model training, as described in Section~\ref{subsec:training_details}. 
    \item 
    MMLU-Medical~\cite{hendrycks2020measuring}: MMLU was originally designed to assess the world knowledge of models across various subjects including mathematics, physics, history, and law. 
    \citet{singhal2023large} created MMLU-Medical by extracting six subjects relevant to the medical field from MMLU, clinical knowledge, medical genetics, anatomy, professional medicine, college biology, and college medicine, aiming to evaluate medical-specialized systems.
    \item 
    JAMA Clinical Challenge~\cite{chen2024benchmarking}: 
    This dataset, sourced from the JAMA Network website,\footnote{\url{jamanetwork.com/collections/44038/clinical-challenge}} comprises long and complex clinical cases. 
    Most questions involve diagnosing or determining the next steps based on the provided case. 
    The dataset adopts a multiple-choice QA format, offering four options for each question.
\end{itemize}

\subsubsection{Completeness and Factuality Evaluation}
The K-QA dataset contains human-generated answers accompanied by detailed statements categorized as ``must-have'' or ``nice to have,'' indicating whether they should be necessarily included in the answer or are additional but beneficial.
Following previous studies~\cite{min2023factscore,manes2024k}, we utilized these atomic facts to employ an evaluation approach based on natural language inference (NLI).
We considered the model's response as the ``premise'' and each human statement as a ``hypothesis.'' 
Using GPT-4, we assessed whether the premise entails or contradicts each hypothesis.\footnote{We used the checkpoint of \texttt{gpt-4-1106-preview}, and we used the evaluation script and input prompt provided by~\citet{manes2024k} (\url{https://github.com/Itaymanes/K-QA}).}
The completeness score $S_\text{comp}$ evaluates how well the model's response entails the must-have statements associated with each question, and the factuality score $S_\text{fact}$ checks if the model's response contradicts any of the must-have or nice-to-have statements as follows:
\begin{equation}
    \begin{aligned}
    & S_\text{comp}(r_i,\mathcal{A}^{\prime}_i) = \sum_{a \in \mathcal{A}^{\prime}_i} \frac{\mathbf{1}[r_i~\text{entails}~a]}{|\mathcal{A}^{\prime}_i|}, \\
    & S_\text{fact}(r_i,\mathcal{A}_i) = \begin{cases} 
        0 & \text{if } \exists a \in \mathcal{A}_i \text{ such that}~r_i~\text{contradicts}~a \\
        1 & \text{otherwise},
    \end{cases}
    \end{aligned}
\end{equation}
where $r_i$ is the model response to the $i$-th question, $\mathcal{A}_i$ is the list of all statements corresponding to the $i$-th questions, $\mathcal{A}^{\prime}_i$ is the list of must-have statements, and $a$ is a statement.
The indicator function $\mathbf{1}[cond]$ returns 1 if the condition is true; otherwise 0.
Both completeness and factuality scores are averaged across all questions in the dataset.

\subsubsection{Details of Baseline Models}

In Figures~\ref{fig:main_results} and~\ref{fig:ablation}, we trained LLaMA-7B and MediTron-7B with a learning rate of 2e-5 for three epochs, while Mistral-7B, BioMistral-7B, and Gemma-7B were trained with a learning rate of 2e-6 for three epochs.
Other hyperparameters and computational environments were similar to those used for Meerkat-7B (described in Section~\ref{sec:training_details}).
We observed significant variations in the models' performance depending on the learning rate and training epoch. Consequently, we selected the better-performing option between the learning rates of 2e-5 and 2e-6, and between three and five epochs.
For LLaMA-3-8B, we did not optimize the hyperparameters, instead using a learning rate of 5e-6 and training for three epochs.
We did not conduct further training on the instruction-tuned models (e.g., ChatDoctor-7B, Med-Alpaca-13B, PMC-LLaMA-13B, etc.). 
Instead, we referred to the hyperparameters recommended by their respective authors and the official GitHub repositories to generate responses.

\subsubsection{Meerkat Inference}
We used the vLLM platform for fast inference~\cite{kwon2023efficient}.
We applied BFloat16 and greedy decoding for the single-model evaluation in the QA tasks.
In the ensemble evaluation, we used a temperature of 0.7, and a repetition penalty of 1.0. We utilized a choice shuffling ensemble technique~\cite{nori2023can}, which involves randomizing the given options before presenting them to the models and subsequently conducting a majority vote to determine the final predictions.
This helps mitigate potential biases in the position of the correct answer~\cite{ko2020look}.


%% file: tables/train_sets.tex
\begin{table}[t]
\captionsetup{width=\textwidth}
\caption{
Statistics of our instruction-tuning datasets. `
`\# Examples'' denotes the number of training examples for each dataset. 
$\dag$ indicates that the dataset is newly constructed in this study. 
The total number of training examples is 460,625.
}
\label{tab:train_sets}
\begin{tabular}{llr}
\toprule
 \textbf{Target Application} & \textbf{Dataset} &  \textbf{\# Examples} \\
\midrule
Multiple-choice QA & MedQA-CoT$^\dag$~\cite{jin2021disease} & 9,308  \\
& MedBooks-CoT-18$^\dag$ & 77,776 \\
& MedMCQA~\cite{pal2022medmcqa} & 182,821 \\
\midrule
Free-form QA & LiveQA~\cite{abacha2017overview} & 633  \\
& MedicationQA~\cite{abacha2019bridging} & 689  \\
& ChatDoctor-cleaned$^\dag$~\cite{li2023chatdoctor} & 112,165  \\
\midrule
Multi-turn Dialogue & 
MedQA-dialog$^\dag$~\cite{jin2021disease} & 4,839  \\
\midrule
Clinical Note Generation & MTS-dialog~\cite{abacha2023empirical} & 1,201 \\
\midrule
Miscellaneous & AlpaCare~\cite{zhang2023alpacare} & 52,002 \\
\bottomrule
\end{tabular}
\end{table}

%% file: figures/framework.tex
\begin{figure*}[t]
\centering
\includegraphics[width=\linewidth]{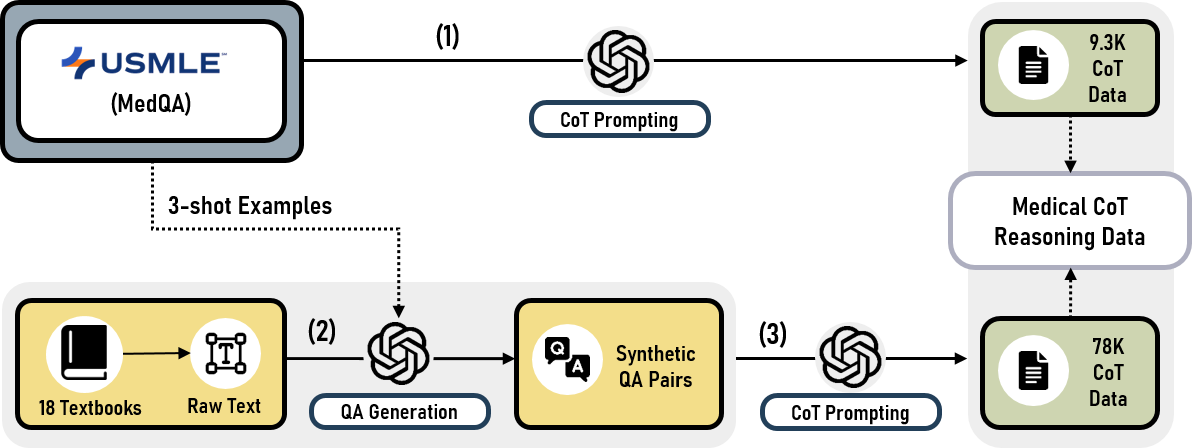} 
\vspace{0.5mm}
\caption{
The overall process of generating synthetic chain-of-thought (CoT) data.
(1) GPT-4 was prompted to provide answers, along with step-by-step explanations, for USMLE-style questions from MedQA~\cite{jin2021disease}, resulting in the creation of 9.3K CoT data.
(2) GPT-4 received three randomly sampled questions from MedQA and text chunks from medical textbooks as input to produce synthetic question-answer pairs.
(3) GPT-4 was then prompted to generate step-by-step explanations for these generated questions, resulting in an additional 78K CoT data.
}
\label{fig:framework}
\end{figure*}

%% file: figures/cot_comparison.tex
\begin{figure*}[t]
\centering
\includegraphics[width=\linewidth]{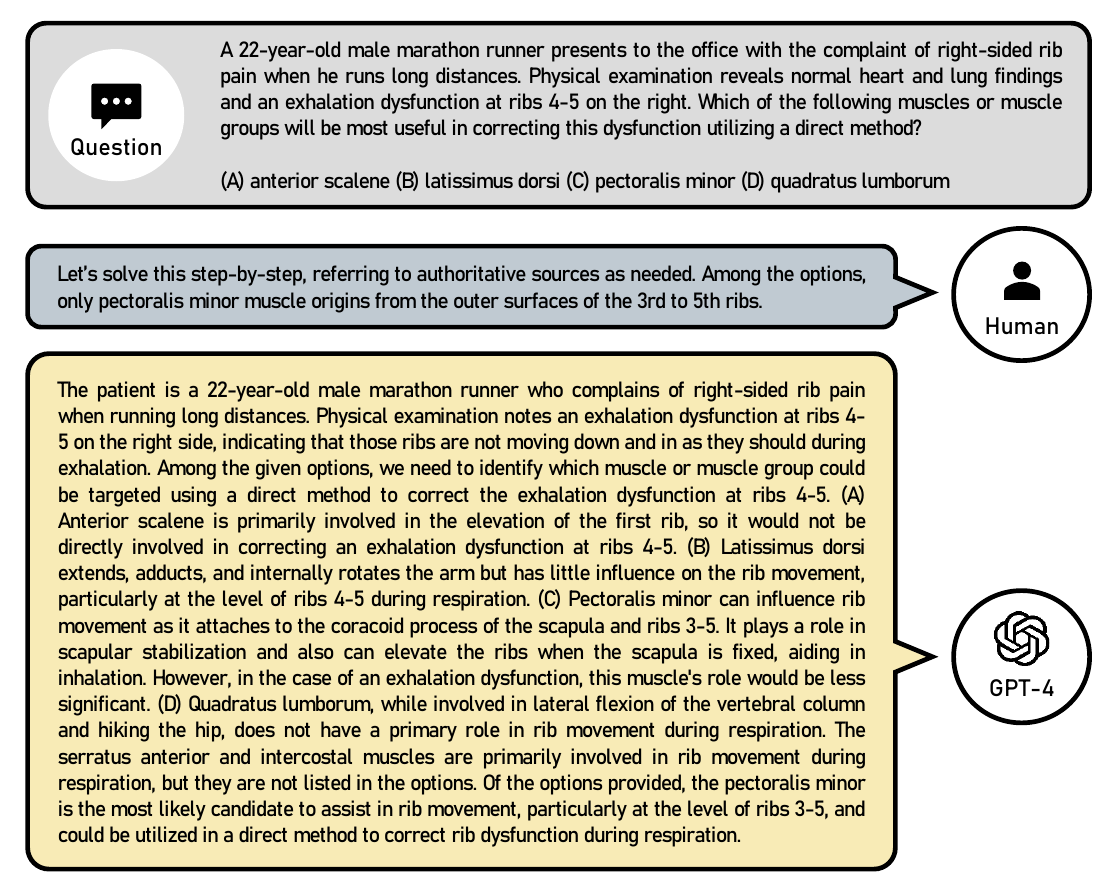} 
\caption{
Examples of explanations provided by human experts and GPT-4 for a USMLE-style question.
GPT-4 offers answers that are notably more detailed, comprehensive, and accurate compared to those from humans. 
The human explanation is sourced from the study of \citet{singhal2023towards}.
}
\label{fig:cot_comparison}
\end{figure*}

%% file: figures/generated_question.tex
\begin{figure*}[t]
\centering
\includegraphics[width=\linewidth]{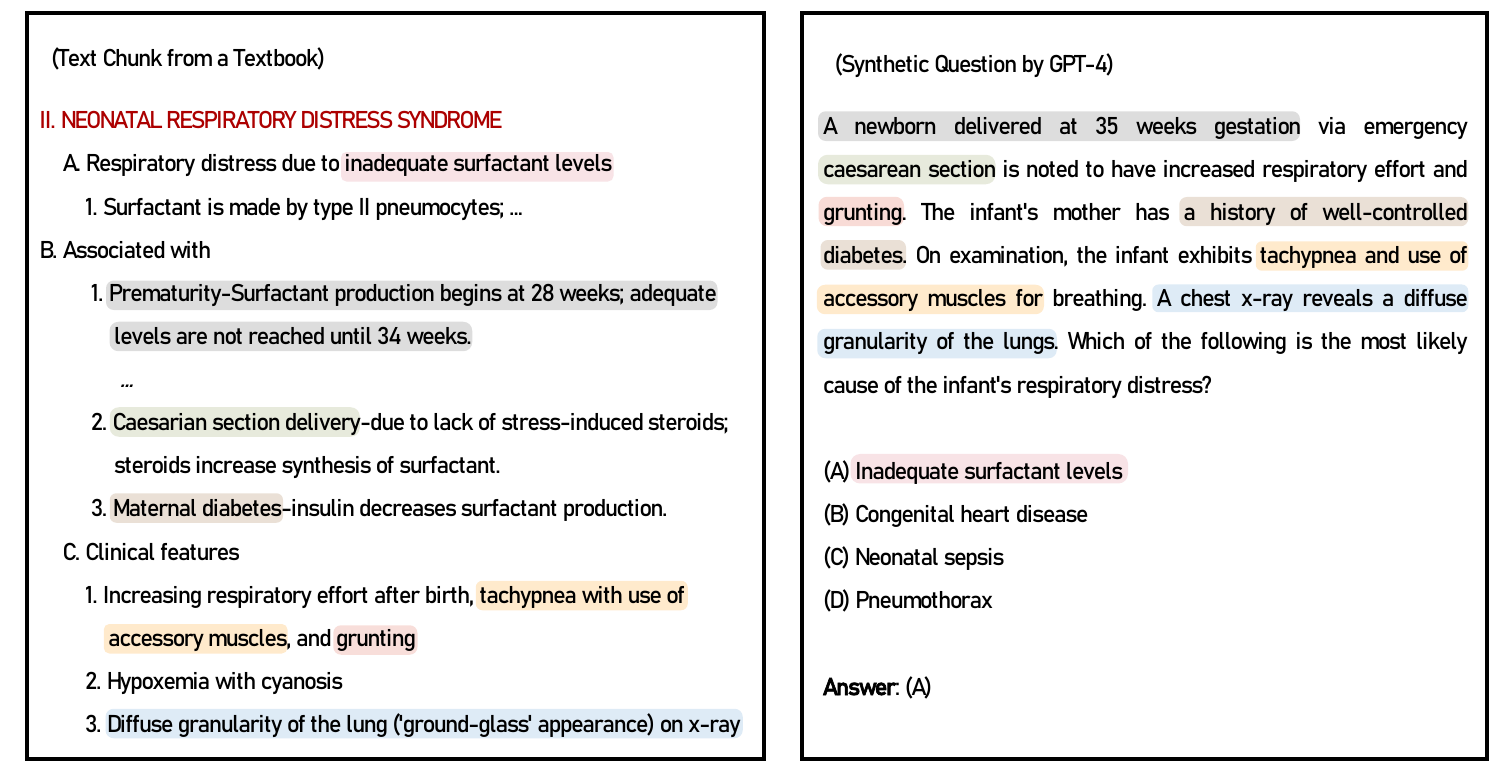} 
\caption{
Example of a question and answer generated from a textbook chunk~\cite{sattar2020fundamentals}.
Spans highlighted in the same color indicate related content. It's evident that GPT-4 heavily relies on the textbook to generate questions, resulting in a significant reduction in hallucination and ensuring that the generated QA examples cover essential content from the textbook. This leads to an overall improvement in quality.
}
\label{fig:generated_question}
\end{figure*}

%% file: tables/test_sets.tex
\begin{table}[t]
\caption{Statistics of benchmark datasets and the evaluation metrics used. 
``\# Examples'': the number of test examples for each dataset. 
The K-QA dataset requires free-form responses as answers, while the other datasets consist of multiple-choice QA problems where the correct answer must be chosen from given options.
}
\label{tab:test_sets}
\begin{tabular}{lllr}
\toprule
 \textbf{Type} & \textbf{Dataset} & \textbf{Metric} & \textbf{\# Examples} \\
\midrule
USMLE & MedQA~\cite{jin2021disease} & Accuracy & 1,273 \\ 
& USMLE sample test & & 325  \\
& Medbullets-4~\cite{chen2024benchmarking} & & 308  \\
& Medbullets-5~\cite{chen2024benchmarking} & & 308  \\
\midrule
Other exams & MedMCQA~\cite{pal2022medmcqa} & Accuracy & 4,182 \\
\cmidrule(lr){2-4}
& MMLU-Medical~\cite{hendrycks2020measuring} & Average Accuracy & 1,089 \\
& - Clinical knowledge & & 265 \\
& - Medical genetics & & 100 \\
& - Anatomy & & 135 \\
& - Professional medicine & & 272 \\
& - College biology & & 144 \\
& - College medicine & & 173 \\
\midrule
Case Challenges & NEJM Case Challenges~\cite{eriksen2023use} & Accuracy & 38 \\
\midrule
Real-world Queries & K-QA~\cite{chen2024benchmarking} &  \begin{tabular}[c]{@{}l@{}}Completeness\\Factuality\end{tabular} & 201 \\
\bottomrule
\end{tabular}
\end{table}

%% file: tables/six_benchmarks.tex
\begin{table}[h!]
\caption{
Performance of models for the six QA benchmarks and six medical subjects in MMLU-Medical~\cite{hendrycks2020measuring}.
The scores in MMLU-Medical were calculated based on the average accuracies across the six medical-related subjects in the original MMLU benchmark.
}\label{tab:six_benchmarks}
\begin{tabular}{lccccccc}
\toprule
\multirow{3}{*}{\textbf{Dataset}} & \multirow{3}{*}{\begin{tabular}[c]{@{}c@{}}\textbf{GPT-4}\\ \textbf{(5-shot)}\end{tabular}} & \multirow{3}{*}{\begin{tabular}[c]{@{}c@{}}\textbf{GPT-3.5}\\ \textbf{(5-shot)}\end{tabular}} & \multirow{3}{*}{\begin{tabular}[c]{@{}c@{}}\textbf{MediTron}\\ \textbf{7B}\end{tabular}} & \multirow{3}{*}{\begin{tabular}[c]{@{}c@{}}\textbf{BioMistral}\\ \textbf{7B}\end{tabular}} & \multicolumn{3}{c}{\textbf{Meerkat}} \\
\cmidrule(lr){6-8}
 & &  &  &  & \textbf{7B} & \textbf{8B} & \textbf{70B} \\
\midrule
Six QA Benchmarks\\
\midrule
MedQA & 81.4 & 53.6 & 50.2 & 54.3 & 70.6 & 74.0 & \underline{82.6} \\
USMLE & 86.6 & 58.5 & 44.6 & 51.4 & 70.3 & 74.2 & \underline{87.4} \\
Medbullets-4 & 68.8 & 51.0 & 51.1 & 52.3 & 58.7 & 62.3 & \underline{71.4} \\
Medbullets-5 & 63.3 & 47.4 & 45.5 & 48.7 & 52.9 & 55.5 & \underline{65.3} \\
MedMCQA & 72.4 & 51.0 & 57.9 & 61.1 & 60.6 & 62.7 & \underline{73.9} \\
MMLU-Medical & \underline{87.1} & 67.3 & 56.7 & 64.6 & 70.5 & 75.2 & 86.9 \\
\midrule
\multicolumn{8}{l}{Six Medical Subjects in MMLU-Medical} \\
\midrule
Clinical Knowledge & 86.4 & 68.7 & 57.7 & 59.9 & 71.6 & 74.3 & \underline{87.2} \\
Medicine Genetics & \underline{92.0} & 68.0 & 63.8 & 64.0 & 74.8 & 76.7 & 88.2 \\
Anatomy & 80.0 & 60.7 & 56.9 & 56.5 & 63.2 & 74.8 & \underline{84.4} \\
Professional Medicine & \underline{93.8} & 69.9 & 56.0 & 60.4 & 77.3 & 75.3 & 87.2 \\
College Biology & \underline{93.8} & 72.9 & 57.1 & 59.0 & 70.8 & 76.1 & 87.9 \\
College Medicine & 76.3 & 63.6 & 48.9 & 54.7 & 65.2 & 74.3 & \underline{86.6} \\
\bottomrule
\end{tabular}
\end{table}

%% file: tables/case_study_longform_1.tex
\begin{table}[t]
\captionsetup{width=\textwidth}
\caption{
Comparison of responses by human, GPT-4, ChatDoctor-7B, and Meerkat-7B for the given question.
ChatDoctor-7B provided an incorrect response, stating that Bactrim is unlikely to cause a yeast infection, which contradicts the human response.
On the other hand, GPT-4 and our Meerkat-7B model correctly stated that Bactrim could potentially cause a yeast infection, providing additional explanations to support their answers.
}\label{tab:case_1}
\begin{tabular}{p{12.5cm}}
\toprule
\textbf{Q}: {Could bactrim potentially cause a yeast infection?} \\
\midrule
\texttt{Human}: The vagina naturally contains populations of yeast, such as candida, and bacteria, such as lactobacillus, that prevent yeast overgrowth. Using antibiotics such as Bactrim will treat the infectious bacteria, but it could coincidentally damage the lactobacillus population, resulting in yeast overgrowth and, therefore, a vaginal yeast infection. \\
\midrule
\texttt{GPT-4}: Yes, Bactrim, which is a brand name for a combination of sulfamethoxazole and trimethoprim, is an antibiotic used to treat various bacterial infections. While it is effective against bacteria, Bactrim, like other antibiotics, does not distinguish between harmful bacteria and the beneficial bacteria that reside in the body, such as those in the gut and the vaginal flora.
\\\\
When antibiotics disrupt the natural balance of bacteria in the body, they can decrease the number of beneficial bacteria that help to keep yeast (Candida) in check. Without sufficient beneficial bacteria, yeast can overgrow, leading to a yeast infection, medically known as candidiasis. This is why some individuals, particularly women, may experience yeast infections following a course of antibiotics, including Bactrim.
\\\\
If you suspect you have developed a yeast infection after taking Bactrim or any other antibiotic, it's important to consult with a healthcare provider. They can provide appropriate diagnosis and treatment options, which may include antifungal medications to restore the natural balance of microorganisms in your body.
\\
\midrule
\texttt{ChatDoctor-7B}:
Bactrim is unlikely to cause a yeast infection as it is an antibiotic and not an anti-fungal medication. \\
\midrule
\texttt{Meerkat-7B}:
Yes, bactrim can potentially cause a yeast infection. Bactrim is an antibiotic that is used to treat bacterial infections. However, it can disrupt the natural balance of bacteria in the body, which can lead to the overgrowth of yeast and result in a yeast infection. It is important to take bactrim as prescribed by your doctor and to seek medical attention if you experience symptoms of a yeast infection, such as itching, redness, and discharge in the vaginal area.
\\
\bottomrule
\end{tabular}
\end{table}

%% file: tables/case_study_longform_2.tex
\begin{table}[t]
\captionsetup{width=\textwidth}
\caption{
Comparison of responses by human, GPT-4, ChatDoctor-7B, and Meerkat-7B for the given question.
The human response highlighted that empiric treatment is appropriate when common symptoms of chlamydia or gonorrhea are present.
GPT-4 effectively distinguished scenarios where empiric treatment is warranted, providing the most precise and detailed explanation among all models.
ChatDoctor-7B contradicted the human response by stating that test results are important without specifying the corresponding scenarios.
Meerkat-7B emphasized the importance of immediate treatment to prevent infection in cases of chlamydia or gonorrhea, implying that waiting for test results may not always be necessary.
}\label{tab:case_2}
\begin{tabular}{p{12.5cm}}
\toprule
\textbf{Q}: Do I need to get test results before starting treatment for either chylamdia or the ghonerrea? \\
\midrule
\texttt{Human}: Chlamydia and gonorrhea are bacteria that can cause sexually transmitted diseases and are treated with antibiotics. Diagnosis for these diseases is performed by laboratory testing of urine, vaginal, rectal, and throat samples. Empiric treatment, meaning treatment without test results, should be offered by the doctor to patients with common symptoms of a sexually transmitted disease, and for those with a recent known or possible sexual exposure to chlamydia or gonorrhea. For patients without common symptoms of a sexually transmitted disease or a suspected exposure, establishing the diagnosis with a positive test result is necessary prior to the initialization of treatment. \\
\midrule
\texttt{GPT-4}: 
For most medical conditions, including sexually transmitted infections (STIs) such as chlamydia and gonorrhea, it is standard practice to obtain an accurate diagnosis before initiating treatment. This typically involves testing to confirm the presence of the infection.
\\\\
... (omitted content) ...
\\\\
However, there may be certain situations where a healthcare provider prescribes treatment before test results are available:
\\\\
**Empiric Treatment**: In some cases, if your symptoms strongly suggest an STI and/or you have had a known exposure to chlamydia or gonorrhea, your healthcare provider might start treatment before test results come back. This is known as empiric treatment and is usually done in the interest of time, to reduce transmission, and to alleviate symptoms more quickly.
\\\\
**Test and Treat**: When access to follow-up care is limited or when there is high concern about patients not returning for treatment if they test positive, some clinics may use a "test and treat" strategy where they test for STIs and treat presumptively at the same time.
\\\\
If you think you may have been exposed to an STI or are experiencing symptoms, it is important to contact a healthcare provider. They can advise you on the appropriate steps to take, including testing and potential treatment. Always follow the advice of your healthcare provider, as they will tailor their recommendations to your specific situation.
\\
\midrule
\texttt{ChatDoctor-7B}: 
Yes, it is important to have any relevant test results before beginning treatment for either condition. \\
\midrule
\texttt{Meerkat-7B}: 
No, you don't need to wait for test results before starting treatment for either chlamydia or gonorrhea. It's important to begin treatment as soon as possible to prevent complications and reduce the risk of spreading the infection to others. Treating these infections is straightforward with antibiotics. The most commonly used antibiotics for chlamydia are azithromycin and doxycycline, while ceftriaxone and azithromycin are commonly used for gonorrhea. Your doctor or healthcare provider will prescribe the appropriate medication based on your specific situation. It's also crucial to ensure that your sexual partner(s) are tested and treated if necessary to completely eliminate the infection and prevent reinfection. If you have any concerns or questions, don't hesitate to discuss them with your healthcare provider.
\\
\bottomrule
\end{tabular}
\end{table}

%% file: tables/case_study_longform_3.tex
\begin{table}[t]
\captionsetup{width=\textwidth}
\caption{
Comparison of responses by human, GPT-4, ChatDoctor-7B, and Meerkat-7B for the given question.
GPT-4 accurately stated that Saxenda dosage titration starts at 0.6 mg and increases by 0.6 mg weekly until reaching 3.0 mg. In contrast, ChatDoctor-7B provided the completely wrong answer ``stay on the same dose.''
While Meerkat-7B generally provided factual responses, it incorrectly mentioned the duration of dose escalation as ``a total of 12 weeks,'' suggesting a need for improvement in delivering more reliable responses.
}\label{tab:case_3}
\begin{tabular}{p{12.5cm}}
\toprule
\textbf{Q}: How long would I be on Saxenda before I could move to the next
dose?\\
\midrule
\texttt{Human}: 
It is recommended that adult patients start with a dose of 0.6 mg per day for one week. The dose is then increased by 0.6 mg each week over the following 4 weeks until the daily maintenance dose of 3 mg is reached. This dosing schedule is used to minimize the risk of gastrointestinal side effects, such as nausea, vomiting, diarrhea, and constipation. If a patient does not tolerate an increased dose during the dosing schedule, dose escalation can be delayed by one extra week. If the 3 mg daily maintenance dose is not tolerated, it is recommended that treatment be discontinued.  \\
\midrule
\texttt{GPT-4}:
Saxenda (liraglutide) is a prescription medication that is used for chronic weight management in adults with obesity or overweight with at least one weight-related condition such as type 2 diabetes, high blood pressure, or high cholesterol. The dose of Saxenda is gradually increased over time to reduce the likelihood of gastrointestinal side effects.
\\\\
The typical dosing schedule for Saxenda is as follows:
\\\\
1. Start with 0.6 mg per day for one week.

2. Increase the dose by 0.6 mg each week until the full maintenance dose of 3.0 mg per day is reached.
\\\\
This means that you would be on each dose for one week before moving up to the next dose. Here is a more detailed breakdown:
\\\\
- Week 1: 0.6 mg per day

- Week 2: 1.2 mg per day

- Week 3: 1.8 mg per day

- Week 4: 2.4 mg per day

- Week 5 and onward: 3.0 mg per day (maintenance dose)
\\\\
It is important to follow your healthcare provider's instructions regarding the dosing schedule. If you experience any adverse effects, especially gastrointestinal ones such as nausea, vomiting, or diarrhea, you should contact your healthcare provider. They may adjust your dosing schedule or provide additional guidance to manage side effects.
\\\\
Always consult your healthcare provider before making any changes to your medication regimen, and use Saxenda exactly as prescribed by your doctor.\\
\midrule
\texttt{ChatDoctor-7B}:
You should stay on the same dose of Saxenda for at least 12 weeks before increasing it. \\
\midrule
\texttt{Meerkat-7B}:
The recommended starting dose for Saxenda is 0.6 mg/day subcutaneously administered once daily. This dose may be increased in weekly increments of 0.6 mg to a maximum dose of 3 mg/day as tolerated, for a total of 12 weeks of dose escalation. The maintenance dose is the dose that is taken for the majority of the treatment period following dose escalation. The maintenance dose may be 0.6 mg/day up to a maximum of 3 mg/day.
\\
\bottomrule
\end{tabular}
\end{table}

%% file: tables/cot_generation_prompt.tex
\begin{table}[t]
\captionsetup{width=\textwidth}
\caption{
Example of the input prompt for generating CoT data from USMLE-style question.
The ``gpt-4-1106-preview'' checkpoint was used.
}\label{tab:prompt_cot_gen}
\begin{tabular}{p{12.5cm}}
\toprule
The following are multiple choice questions about medical knowledge.
Solve them in a step-by-step fashion, starting by summarizing the available information. 
Output your explanation and single option from the given options as the final answer. 
You are required to use the phrase ``the answer is'' at the end of your response. 
See the below example.
\\\\
\#\# Answer: \{\{your step-by-step explanation here\}\}. Therefore, the answer is (D) \{\{answer string\}\}.
\\\\
This is the input question to solve.
\\\\
\#\# Question: \texttt{[INPUT QUESTION]}
\\\\
\#\# Answer:
\\
\bottomrule
\end{tabular}
\end{table}

%% file: tables/qa_generation_prompt.tex
\begin{table}[t]
\captionsetup{width=\textwidth}
\caption{
Example of the input prompt for generating synthetic QA data from textbooks.
The ``gpt-4-1106-preview'' checkpoint was used.
}\label{tab:prompt_qa_gen}
\begin{tabular}{p{12.5cm}}
\toprule
You are an expert in writing questions on the U.S. medical licensing exam. We will provide you with a reference material excerpted from a textbook. 
You should extract essential medical concepts (i.e., named entities, phrases, quantities, sentences, or their combinations) potentially relevant to the exam from the reference material and use them as the correct answer by modifying or combining them appropriately. 
\\\\
Subsequently, you should write a case representation--a brief description of a patient (e.g., demographic information, signs/symptoms, conditions)--along with a relevant question alined with the correct answer. Each question should be accompanied by a correct answer and three candidate answers, all of which should be connected to the correct answer and carefully designed to be plausible, while not demanding excessively specific medical knowledge. Additionally, it is essential that questions should not be readily answered solely by referencing the given context but answerable with an explanation derived from authoritative sources such as textbooks, guidelines, and journal articles. It is also important to note that the reference, being a section of the textbook, may not always offer useful or pertinent information. In such cases, do not formulate a question.
\\\\
We provide three examples below for better understanding.
\\\\
``question and options'': \texttt{[MEDQA\_QUESTION\_1]}
\\
``answer'': \texttt{[ANSWER\_1]}
\\\\
``question and options'': \texttt{[MEDQA\_QUESTION\_2]}
\\
``answer'': \texttt{[ANSWER\_2]}
\\\\
``question and options'': \texttt{[MEDQA\_QUESTION\_3]}
\\
``answer'': \texttt{[ANSWER\_3]}
\\\\
You are required to produce either ``N/A'' or a JSON without including any other responses. If the reference does not contain a suitable medical concept, respond with ``N/A.'' Otherwise, the output should be organized following the JSON format (write only ONE question):
\\\\
\texttt{json}\{``question'': QUESTION,
``options'': \{``A'': OPT, ``B'': OPT, ...\}, 
``explanation'': \{your\_explanation\}, 
``answer'': ANSWER\}
\\\\
This is the input reference: 
\\
\texttt{[TEXTBOOK CHUNK]}
\\
\bottomrule
\end{tabular}
\end{table}

%% file: tables/chatdoctor_cleaning_prompt.tex
\begin{table}[t]
\captionsetup{width=\textwidth}
\caption{
Example of the input prompt for generating the ChatDoctor-cleaned data.
The ``gpt-3.5-turbo-0613'' checkpoint was used.
}\label{tab:prompt_chatdoctor_cleaning}
\begin{tabular}{p{12.5cm}}
\toprule
Your task is to refine responses to patient inquiries gathered from an online medical consultation platform to mirror the natural communication style of a practicing healthcare expert. You also should remove noise from responses, including patients' or doctors' names, unintended text generated during data crawling, and mentions of specific platforms (e.g., Chat Doctor). 
Refer to the following examples.
\\\\
Original text: Dear Friend. Welcome to Chat Doctor. I am Chat Doctor. I understand your concern. Recurring yeast / final infection occur due to:1. Improper treatment.2. Inadequate treatment3. Wrong diagnosis.4. Not treating the partner. I would advise you to get a Urine examination done for bacteria and fungus. When the reports are available, get treated. Also, treat your partner. This is my personal opinion for you based on available details. If you have any further query please ask me. Stay HealthyChatDoctor, MD
\\\\
Modified text: I understand your concern regarding recurrent yeast/fungal infections. Several factors can contribute to this issue, including improper or inadequate treatment, misdiagnosis, or failure to treat the partner. I recommend getting a urine examination done to check for bacteria and fungus. Once the reports are available, appropriate treatment can be initiated. It's also important to ensure that your partner receives treatment if necessary. Please note that this advice is based on the information provided. If you have any further questions or concerns, feel free to ask. Take care and stay healthy.
\\\\
Original text: Hi, dairy have gone through your question. I can understand your concern.... She has well differentiated keratinizing squamous cell carcinoma.  If her general health is good then treatment of choice is wide excision of carcinoma followed by chemotherapy or radiotherapy if needed. Consult your doctor and take treatment accordingly. Hope I have answered your question, if you have doubt then I will be happy to answer. Thanks for using Chat Doctor. Wish you a very good health.
\\\\
Modified text: I've reviewed your inquiry and understand your concern. It appears she has well-differentiated keratinizing squamous cell carcinoma. If her overall health is good, the preferred treatment is wide excision of the carcinoma, followed by chemotherapy or radiotherapy if necessary. I recommend consulting with your doctor to discuss treatment options and proceed accordingly. If you have any further questions or doubts, feel free to ask. Wishing you good health.
\\\\
Original text: Thank you for using Chat Doctor. I would suggest that you see your doctor. Your baby maybe having bronchiolitis which is a lung infection common to your kids age. It is commonly caused by a virus. Albuterol via nebulization should be utilized in order to alleviate the wheezing and also help with the congestion. A decongestant can also be used for the colds. Also, it would also advise doing a chest X-ray in order to rule out other diseases (ex. pneumonia)sincerely, Mark RosarioGeneral pediatrics/Pediatric Pulmonology
\\\\
Modified text: It sounds like your baby might be experiencing bronchiolitis, a common lung infection among children of their age, typically caused by a virus. To help alleviate the wheezing and congestion, I recommend using Albuterol through nebulization. Additionally, a decongestant can be beneficial for managing cold symptoms. Considering the situation, it would be prudent to consult with your doctor for further evaluation. They may suggest a chest X-ray to rule out other potential conditions, such as pneumonia. Take care.
\\\\
Here's the target text: \texttt{[ORIGINAL RESPONSE]}
\\\\
*Please make sure to thoroughly remove all mentions of ``Chat Doctor'' and person names (e.g., Dear John) from the target text. Also, expressions like ``thank you for your query'' and ``thank you for consulting with us'' are awkward in genuine patient-doctor conversations because these expressions are more appropriate on online platforms; therefore, just remove them or replace them with different expressions that feel more natural and convey empathy. Your response:
\\
\bottomrule
\end{tabular}
\end{table}

%% file: tables/dialogue_generation_prompt.tex
\begin{table}[t]
\captionsetup{width=\textwidth}
\caption{
Example of the input prompt for generating the MedQA-dialogue data.
The ``gpt-3.5-turbo-0613'' checkpoint was used. 
}\label{tab:prompt_dialogue_gen}
\begin{tabular}{p{12.5cm}}
\toprule
Your task is to transform a provided question and answer into a dialogue between a patient, a patient's guardian, student, or researcher (hereinafter referred to ``Person A'') and a doctor or healthcare professional (hereinafter referred to ``Person B''). 
Person A will communicate their symptoms, medical history, and, if relevant, demographic details. 
Person B will guide the conversation, probing for additional information when necessary. 
You have the option to compose a concise, single-turn conversation if the patient's input is comprehensive. 
However, if essential details are missing, Person B should engage in a multi-turn dialogue, asking follow-up questions to gather a thorough medical history and records. 
The presumption is that Person B possesses no prior information about Person A, including vital signs such as temperature and blood pressure, until Person A discloses it during the conversation. 
Additionally, Person B must prioritize conveying crucial information about the correct option. 
Person B can additionally discuss other options if Person A explicitly inquires about them. 
Person A does not need to provide an introduction and his/her name to Person B. 
Person B can lead the conversation without feeling the need to respond to every statement from Person A. 
The dialogue always starts with Person A and ends with Person B.
\\\\
Here's the input question and answer.
\\\\
Question: \texttt{[MEDQA QUESTION]}
\\\\
Answer: \texttt{[ANSWER WITH COT]}
\\
\bottomrule
\end{tabular}
\end{table}

%% file: main.bbl

\begin{thebibliography}{53}
\ifx \bisbn   \undefined \def \bisbn  #1{ISBN #1}\fi
\ifx \binits  \undefined \def \binits#1{#1}\fi
\ifx \bauthor  \undefined \def \bauthor#1{#1}\fi
\ifx \batitle  \undefined \def \batitle#1{#1}\fi
\ifx \bjtitle  \undefined \def \bjtitle#1{#1}\fi
\ifx \bvolume  \undefined \def \bvolume#1{\textbf{#1}}\fi
\ifx \byear  \undefined \def \byear#1{#1}\fi
\ifx \bissue  \undefined \def \bissue#1{#1}\fi
\ifx \bfpage  \undefined \def \bfpage#1{#1}\fi
\ifx \blpage  \undefined \def \blpage #1{#1}\fi
\ifx \burl  \undefined \def \burl#1{\textsf{#1}}\fi
\ifx \doiurl  \undefined \def \doiurl#1{\url{https://doi.org/#1}}\fi
\ifx \betal  \undefined \def \betal{\textit{et al.}}\fi
\ifx \binstitute  \undefined \def \binstitute#1{#1}\fi
\ifx \binstitutionaled  \undefined \def \binstitutionaled#1{#1}\fi
\ifx \bctitle  \undefined \def \bctitle#1{#1}\fi
\ifx \beditor  \undefined \def \beditor#1{#1}\fi
\ifx \bpublisher  \undefined \def \bpublisher#1{#1}\fi
\ifx \bbtitle  \undefined \def \bbtitle#1{#1}\fi
\ifx \bedition  \undefined \def \bedition#1{#1}\fi
\ifx \bseriesno  \undefined \def \bseriesno#1{#1}\fi
\ifx \blocation  \undefined \def \blocation#1{#1}\fi
\ifx \bsertitle  \undefined \def \bsertitle#1{#1}\fi
\ifx \bsnm \undefined \def \bsnm#1{#1}\fi
\ifx \bsuffix \undefined \def \bsuffix#1{#1}\fi
\ifx \bparticle \undefined \def \bparticle#1{#1}\fi
\ifx \barticle \undefined \def \barticle#1{#1}\fi
\bibcommenthead
\ifx \bconfdate \undefined \def \bconfdate #1{#1}\fi
\ifx \botherref \undefined \def \botherref #1{#1}\fi
\ifx \url \undefined \def \url#1{\textsf{#1}}\fi
\ifx \bchapter \undefined \def \bchapter#1{#1}\fi
\ifx \bbook \undefined \def \bbook#1{#1}\fi
\ifx \bcomment \undefined \def \bcomment#1{#1}\fi
\ifx \oauthor \undefined \def \oauthor#1{#1}\fi
\ifx \citeauthoryear \undefined \def \citeauthoryear#1{#1}\fi
\ifx \endbibitem  \undefined \def \endbibitem {}\fi
\ifx \bconflocation  \undefined \def \bconflocation#1{#1}\fi
\ifx \arxivurl  \undefined \def \arxivurl#1{\textsf{#1}}\fi
\csname PreBibitemsHook\endcsname

\bibitem[\protect\citeauthoryear{Thirunavukarasu et~al.}{2023}]{thirunavukarasu2023large}
\begin{barticle}
\bauthor{\bsnm{Thirunavukarasu}, \binits{A.J.}},
\bauthor{\bsnm{Ting}, \binits{D.S.J.}},
\bauthor{\bsnm{Elangovan}, \binits{K.}},
\bauthor{\bsnm{Gutierrez}, \binits{L.}},
\bauthor{\bsnm{Tan}, \binits{T.F.}},
\bauthor{\bsnm{Ting}, \binits{D.S.W.}}:
\batitle{Large language models in medicine}.
\bjtitle{Nature medicine}
\bvolume{29}(\bissue{8}),
\bfpage{1930}--\blpage{1940}
(\byear{2023})
\end{barticle}
\endbibitem

\bibitem[\protect\citeauthoryear{Tian et~al.}{2024}]{tian2024opportunities}
\begin{barticle}
\bauthor{\bsnm{Tian}, \binits{S.}},
\bauthor{\bsnm{Jin}, \binits{Q.}},
\bauthor{\bsnm{Yeganova}, \binits{L.}},
\bauthor{\bsnm{Lai}, \binits{P.-T.}},
\bauthor{\bsnm{Zhu}, \binits{Q.}},
\bauthor{\bsnm{Chen}, \binits{X.}},
\bauthor{\bsnm{Yang}, \binits{Y.}},
\bauthor{\bsnm{Chen}, \binits{Q.}},
\bauthor{\bsnm{Kim}, \binits{W.}},
\bauthor{\bsnm{Comeau}, \binits{D.C.}}, \betal:
\batitle{Opportunities and challenges for chatgpt and large language models in biomedicine and health}.
\bjtitle{Briefings in Bioinformatics}
\bvolume{25}(\bissue{1}),
\bfpage{493}
(\byear{2024})
\end{barticle}
\endbibitem

\bibitem[\protect\citeauthoryear{Kung et~al.}{2023}]{kung2023performance}
\begin{barticle}
\bauthor{\bsnm{Kung}, \binits{T.H.}},
\bauthor{\bsnm{Cheatham}, \binits{M.}},
\bauthor{\bsnm{Medenilla}, \binits{A.}},
\bauthor{\bsnm{Sillos}, \binits{C.}},
\bauthor{\bsnm{De~Leon}, \binits{L.}},
\bauthor{\bsnm{Elepa{\~n}o}, \binits{C.}},
\bauthor{\bsnm{Madriaga}, \binits{M.}},
\bauthor{\bsnm{Aggabao}, \binits{R.}},
\bauthor{\bsnm{Diaz-Candido}, \binits{G.}},
\bauthor{\bsnm{Maningo}, \binits{J.}}, \betal:
\batitle{Performance of chatgpt on usmle: Potential for ai-assisted medical education using large language models}.
\bjtitle{PLoS digital health}
\bvolume{2}(\bissue{2}),
\bfpage{0000198}
(\byear{2023})
\end{barticle}
\endbibitem

\bibitem[\protect\citeauthoryear{Singhal et~al.}{2023a}]{singhal2023large}
\begin{barticle}
\bauthor{\bsnm{Singhal}, \binits{K.}},
\bauthor{\bsnm{Azizi}, \binits{S.}},
\bauthor{\bsnm{Tu}, \binits{T.}},
\bauthor{\bsnm{Mahdavi}, \binits{S.S.}},
\bauthor{\bsnm{Wei}, \binits{J.}},
\bauthor{\bsnm{Chung}, \binits{H.W.}},
\bauthor{\bsnm{Scales}, \binits{N.}},
\bauthor{\bsnm{Tanwani}, \binits{A.}},
\bauthor{\bsnm{Cole-Lewis}, \binits{H.}},
\bauthor{\bsnm{Pfohl}, \binits{S.}}, \betal:
\batitle{Large language models encode clinical knowledge}.
\bjtitle{Nature}
\bvolume{620}(\bissue{7972}),
\bfpage{172}--\blpage{180}
(\byear{2023})
\end{barticle}
\endbibitem

\bibitem[\protect\citeauthoryear{Singhal et~al.}{2023b}]{singhal2023towards}
\begin{botherref}
\oauthor{\bsnm{Singhal}, \binits{K.}},
\oauthor{\bsnm{Tu}, \binits{T.}},
\oauthor{\bsnm{Gottweis}, \binits{J.}},
\oauthor{\bsnm{Sayres}, \binits{R.}},
\oauthor{\bsnm{Wulczyn}, \binits{E.}},
\oauthor{\bsnm{Hou}, \binits{L.}},
\oauthor{\bsnm{Clark}, \binits{K.}},
\oauthor{\bsnm{Pfohl}, \binits{S.}},
\oauthor{\bsnm{Cole-Lewis}, \binits{H.}},
\oauthor{\bsnm{Neal}, \binits{D.}}, et al.:
Towards expert-level medical question answering with large language models.
arXiv preprint arXiv:2305.09617
(2023)
\end{botherref}
\endbibitem

\bibitem[\protect\citeauthoryear{Nori et~al.}{2023}]{nori2023capabilities}
\begin{botherref}
\oauthor{\bsnm{Nori}, \binits{H.}},
\oauthor{\bsnm{King}, \binits{N.}},
\oauthor{\bsnm{McKinney}, \binits{S.M.}},
\oauthor{\bsnm{Carignan}, \binits{D.}},
\oauthor{\bsnm{Horvitz}, \binits{E.}}:
Capabilities of gpt-4 on medical challenge problems.
arXiv preprint arXiv:2303.13375
(2023)
\end{botherref}
\endbibitem

\bibitem[\protect\citeauthoryear{Brin et~al.}{2023}]{brin2023comparing}
\begin{barticle}
\bauthor{\bsnm{Brin}, \binits{D.}},
\bauthor{\bsnm{Sorin}, \binits{V.}},
\bauthor{\bsnm{Vaid}, \binits{A.}},
\bauthor{\bsnm{Soroush}, \binits{A.}},
\bauthor{\bsnm{Glicksberg}, \binits{B.S.}},
\bauthor{\bsnm{Charney}, \binits{A.W.}},
\bauthor{\bsnm{Nadkarni}, \binits{G.}},
\bauthor{\bsnm{Klang}, \binits{E.}}:
\batitle{Comparing chatgpt and gpt-4 performance in usmle soft skill assessments}.
\bjtitle{Scientific Reports}
\bvolume{13}(\bissue{1}),
\bfpage{16492}
(\byear{2023})
\end{barticle}
\endbibitem

\bibitem[\protect\citeauthoryear{Saab et~al.}{2024}]{saab2024capabilities}
\begin{botherref}
\oauthor{\bsnm{Saab}, \binits{K.}},
\oauthor{\bsnm{Tu}, \binits{T.}},
\oauthor{\bsnm{Weng}, \binits{W.-H.}},
\oauthor{\bsnm{Tanno}, \binits{R.}},
\oauthor{\bsnm{Stutz}, \binits{D.}},
\oauthor{\bsnm{Wulczyn}, \binits{E.}},
\oauthor{\bsnm{Zhang}, \binits{F.}},
\oauthor{\bsnm{Strother}, \binits{T.}},
\oauthor{\bsnm{Park}, \binits{C.}},
\oauthor{\bsnm{Vedadi}, \binits{E.}}, et al.:
Capabilities of gemini models in medicine.
arXiv preprint arXiv:2404.18416
(2024)
\end{botherref}
\endbibitem

\bibitem[\protect\citeauthoryear{Zakka et~al.}{2024}]{zakka2024almanac}
\begin{barticle}
\bauthor{\bsnm{Zakka}, \binits{C.}},
\bauthor{\bsnm{Shad}, \binits{R.}},
\bauthor{\bsnm{Chaurasia}, \binits{A.}},
\bauthor{\bsnm{Dalal}, \binits{A.R.}},
\bauthor{\bsnm{Kim}, \binits{J.L.}},
\bauthor{\bsnm{Moor}, \binits{M.}},
\bauthor{\bsnm{Fong}, \binits{R.}},
\bauthor{\bsnm{Phillips}, \binits{C.}},
\bauthor{\bsnm{Alexander}, \binits{K.}},
\bauthor{\bsnm{Ashley}, \binits{E.}}, \betal:
\batitle{Almanac—retrieval-augmented language models for clinical medicine}.
\bjtitle{NEJM AI}
\bvolume{1}(\bissue{2}),
\bfpage{2300068}
(\byear{2024})
\end{barticle}
\endbibitem

\bibitem[\protect\citeauthoryear{Tu et~al.}{2024}]{tu2024towards}
\begin{botherref}
\oauthor{\bsnm{Tu}, \binits{T.}},
\oauthor{\bsnm{Palepu}, \binits{A.}},
\oauthor{\bsnm{Schaekermann}, \binits{M.}},
\oauthor{\bsnm{Saab}, \binits{K.}},
\oauthor{\bsnm{Freyberg}, \binits{J.}},
\oauthor{\bsnm{Tanno}, \binits{R.}},
\oauthor{\bsnm{Wang}, \binits{A.}},
\oauthor{\bsnm{Li}, \binits{B.}},
\oauthor{\bsnm{Amin}, \binits{M.}},
\oauthor{\bsnm{Tomasev}, \binits{N.}}, et al.:
Towards conversational diagnostic ai.
arXiv preprint arXiv:2401.05654
(2024)
\end{botherref}
\endbibitem

\bibitem[\protect\citeauthoryear{Eriksen et~al.}{2023}]{eriksen2023use}
\begin{botherref}
\oauthor{\bsnm{Eriksen}, \binits{A.V.}},
\oauthor{\bsnm{M{\"o}ller}, \binits{S.}},
\oauthor{\bsnm{Ryg}, \binits{J.}}:
Use of GPT-4 to diagnose complex clinical cases.
Massachusetts Medical Society
(2023)
\end{botherref}
\endbibitem

\bibitem[\protect\citeauthoryear{Li and Zhang}{2017}]{li2017exploration}
\begin{bchapter}
\bauthor{\bsnm{Li}, \binits{X.}},
\bauthor{\bsnm{Zhang}, \binits{T.}}:
\bctitle{An exploration on artificial intelligence application: From security, privacy and ethic perspective}.
In: \bbtitle{2017 IEEE 2nd International Conference on Cloud Computing and Big Data Analysis (ICCCBDA)},
pp. \bfpage{416}--\blpage{420}
(\byear{2017}).
\bcomment{IEEE}
\end{bchapter}
\endbibitem

\bibitem[\protect\citeauthoryear{Bartoletti}{2019}]{bartoletti2019ai}
\begin{bchapter}
\bauthor{\bsnm{Bartoletti}, \binits{I.}}:
\bctitle{Ai in healthcare: Ethical and privacy challenges}.
In: \bbtitle{Artificial Intelligence in Medicine: 17th Conference on Artificial Intelligence in Medicine, AIME 2019, Poznan, Poland, June 26--29, 2019, Proceedings 17},
pp. \bfpage{7}--\blpage{10}
(\byear{2019}).
\bcomment{Springer}
\end{bchapter}
\endbibitem

\bibitem[\protect\citeauthoryear{Mesk{\'o} and Topol}{2023}]{mesko2023imperative}
\begin{barticle}
\bauthor{\bsnm{Mesk{\'o}}, \binits{B.}},
\bauthor{\bsnm{Topol}, \binits{E.J.}}:
\batitle{The imperative for regulatory oversight of large language models (or generative ai) in healthcare}.
\bjtitle{NPJ digital medicine}
\bvolume{6}(\bissue{1}),
\bfpage{120}
(\byear{2023})
\end{barticle}
\endbibitem

\bibitem[\protect\citeauthoryear{OpenAI}{2022}]{openai2022chatgpt}
\begin{botherref}
\oauthor{\bsnm{OpenAI}}:
Introducing chatgpt
(2022)
\end{botherref}
\endbibitem

\bibitem[\protect\citeauthoryear{Achiam et~al.}{2023}]{achiam2023gpt}
\begin{botherref}
\oauthor{\bsnm{Achiam}, \binits{J.}},
\oauthor{\bsnm{Adler}, \binits{S.}},
\oauthor{\bsnm{Agarwal}, \binits{S.}},
\oauthor{\bsnm{Ahmad}, \binits{L.}},
\oauthor{\bsnm{Akkaya}, \binits{I.}},
\oauthor{\bsnm{Aleman}, \binits{F.L.}},
\oauthor{\bsnm{Almeida}, \binits{D.}},
\oauthor{\bsnm{Altenschmidt}, \binits{J.}},
\oauthor{\bsnm{Altman}, \binits{S.}},
\oauthor{\bsnm{Anadkat}, \binits{S.}}, et al.:
Gpt-4 technical report.
arXiv preprint arXiv:2303.08774
(2023)
\end{botherref}
\endbibitem

\bibitem[\protect\citeauthoryear{Touvron et~al.}{2023}]{touvron2023llama}
\begin{botherref}
\oauthor{\bsnm{Touvron}, \binits{H.}},
\oauthor{\bsnm{Lavril}, \binits{T.}},
\oauthor{\bsnm{Izacard}, \binits{G.}},
\oauthor{\bsnm{Martinet}, \binits{X.}},
\oauthor{\bsnm{Lachaux}, \binits{M.-A.}},
\oauthor{\bsnm{Lacroix}, \binits{T.}},
\oauthor{\bsnm{Rozi{\`e}re}, \binits{B.}},
\oauthor{\bsnm{Goyal}, \binits{N.}},
\oauthor{\bsnm{Hambro}, \binits{E.}},
\oauthor{\bsnm{Azhar}, \binits{F.}}, et al.:
Llama: Open and efficient foundation language models.
arXiv preprint arXiv:2302.13971
(2023)
\end{botherref}
\endbibitem

\bibitem[\protect\citeauthoryear{Jiang et~al.}{2023}]{jiang2023mistral}
\begin{botherref}
\oauthor{\bsnm{Jiang}, \binits{A.Q.}},
\oauthor{\bsnm{Sablayrolles}, \binits{A.}},
\oauthor{\bsnm{Mensch}, \binits{A.}},
\oauthor{\bsnm{Bamford}, \binits{C.}},
\oauthor{\bsnm{Chaplot}, \binits{D.S.}},
\oauthor{\bsnm{Casas}, \binits{D.d.l.}},
\oauthor{\bsnm{Bressand}, \binits{F.}},
\oauthor{\bsnm{Lengyel}, \binits{G.}},
\oauthor{\bsnm{Lample}, \binits{G.}},
\oauthor{\bsnm{Saulnier}, \binits{L.}}, et al.:
Mistral 7b.
arXiv preprint arXiv:2310.06825
(2023)
\end{botherref}
\endbibitem

\bibitem[\protect\citeauthoryear{Google}{2024}]{google2024gemma}
\begin{botherref}
\oauthor{\bsnm{Google}}:
Gemma: Introducing new state-of-the-art open models
(2024)
\end{botherref}
\endbibitem

\bibitem[\protect\citeauthoryear{Wu et~al.}{2023}]{wu2023pmc}
\begin{botherref}
\oauthor{\bsnm{Wu}, \binits{C.}},
\oauthor{\bsnm{Zhang}, \binits{X.}},
\oauthor{\bsnm{Zhang}, \binits{Y.}},
\oauthor{\bsnm{Wang}, \binits{Y.}},
\oauthor{\bsnm{Xie}, \binits{W.}}:
Pmc-llama: Further finetuning llama on medical papers.
arXiv preprint arXiv:2304.14454
(2023)
\end{botherref}
\endbibitem

\bibitem[\protect\citeauthoryear{Chen et~al.}{2023}]{chen2023meditron}
\begin{botherref}
\oauthor{\bsnm{Chen}, \binits{Z.}},
\oauthor{\bsnm{Cano}, \binits{A.H.}},
\oauthor{\bsnm{Romanou}, \binits{A.}},
\oauthor{\bsnm{Bonnet}, \binits{A.}},
\oauthor{\bsnm{Matoba}, \binits{K.}},
\oauthor{\bsnm{Salvi}, \binits{F.}},
\oauthor{\bsnm{Pagliardini}, \binits{M.}},
\oauthor{\bsnm{Fan}, \binits{S.}},
\oauthor{\bsnm{K{\"o}pf}, \binits{A.}},
\oauthor{\bsnm{Mohtashami}, \binits{A.}}, et al.:
Meditron-70b: Scaling medical pretraining for large language models.
arXiv preprint arXiv:2311.16079
(2023)
\end{botherref}
\endbibitem

\bibitem[\protect\citeauthoryear{Labrak et~al.}{2024}]{labrak2024biomistral}
\begin{botherref}
\oauthor{\bsnm{Labrak}, \binits{Y.}},
\oauthor{\bsnm{Bazoge}, \binits{A.}},
\oauthor{\bsnm{Morin}, \binits{E.}},
\oauthor{\bsnm{Gourraud}, \binits{P.-A.}},
\oauthor{\bsnm{Rouvier}, \binits{M.}},
\oauthor{\bsnm{Dufour}, \binits{R.}}:
Biomistral: A collection of open-source pretrained large language models for medical domains.
arXiv preprint arXiv:2402.10373
(2024)
\end{botherref}
\endbibitem

\bibitem[\protect\citeauthoryear{Xie et~al.}{2024}]{xie2024me}
\begin{botherref}
\oauthor{\bsnm{Xie}, \binits{Q.}},
\oauthor{\bsnm{Chen}, \binits{Q.}},
\oauthor{\bsnm{Chen}, \binits{A.}},
\oauthor{\bsnm{Peng}, \binits{C.}},
\oauthor{\bsnm{Hu}, \binits{Y.}},
\oauthor{\bsnm{Lin}, \binits{F.}},
\oauthor{\bsnm{Peng}, \binits{X.}},
\oauthor{\bsnm{Huang}, \binits{J.}},
\oauthor{\bsnm{Zhang}, \binits{J.}},
\oauthor{\bsnm{Keloth}, \binits{V.}}, et al.:
Me llama: Foundation large language models for medical applications.
arXiv preprint arXiv:2402.12749
(2024)
\end{botherref}
\endbibitem

\bibitem[\protect\citeauthoryear{Wei et~al.}{2022a}]{wei2022chain}
\begin{barticle}
\bauthor{\bsnm{Wei}, \binits{J.}},
\bauthor{\bsnm{Wang}, \binits{X.}},
\bauthor{\bsnm{Schuurmans}, \binits{D.}},
\bauthor{\bsnm{Bosma}, \binits{M.}},
\bauthor{\bsnm{Xia}, \binits{F.}},
\bauthor{\bsnm{Chi}, \binits{E.}},
\bauthor{\bsnm{Le}, \binits{Q.V.}},
\bauthor{\bsnm{Zhou}, \binits{D.}}, \betal:
\batitle{Chain-of-thought prompting elicits reasoning in large language models}.
\bjtitle{Advances in Neural Information Processing Systems}
\bvolume{35},
\bfpage{24824}--\blpage{24837}
(\byear{2022})
\end{barticle}
\endbibitem

\bibitem[\protect\citeauthoryear{Wei et~al.}{2022b}]{wei2022emergent}
\begin{botherref}
\oauthor{\bsnm{Wei}, \binits{J.}},
\oauthor{\bsnm{Tay}, \binits{Y.}},
\oauthor{\bsnm{Bommasani}, \binits{R.}},
\oauthor{\bsnm{Raffel}, \binits{C.}},
\oauthor{\bsnm{Zoph}, \binits{B.}},
\oauthor{\bsnm{Borgeaud}, \binits{S.}},
\oauthor{\bsnm{Yogatama}, \binits{D.}},
\oauthor{\bsnm{Bosma}, \binits{M.}},
\oauthor{\bsnm{Zhou}, \binits{D.}},
\oauthor{\bsnm{Metzler}, \binits{D.}},
\oauthor{\bsnm{Chi}, \binits{E.H.}},
\oauthor{\bsnm{Hashimoto}, \binits{T.}},
\oauthor{\bsnm{Vinyals}, \binits{O.}},
\oauthor{\bsnm{Liang}, \binits{P.}},
\oauthor{\bsnm{Dean}, \binits{J.}},
\oauthor{\bsnm{Fedus}, \binits{W.}}:
Emergent abilities of large language models.
Transactions on Machine Learning Research
(2022)
\end{botherref}
\endbibitem

\bibitem[\protect\citeauthoryear{Tay et~al.}{2022}]{tay2022unifying}
\begin{botherref}
\oauthor{\bsnm{Tay}, \binits{Y.}},
\oauthor{\bsnm{Dehghani}, \binits{M.}},
\oauthor{\bsnm{Tran}, \binits{V.Q.}},
\oauthor{\bsnm{Garcia}, \binits{X.}},
\oauthor{\bsnm{Bahri}, \binits{D.}},
\oauthor{\bsnm{Schuster}, \binits{T.}},
\oauthor{\bsnm{Zheng}, \binits{H.S.}},
\oauthor{\bsnm{Houlsby}, \binits{N.}},
\oauthor{\bsnm{Metzler}, \binits{D.}}:
Unifying language learning paradigms.
arXiv preprint arXiv:2205.05131
(2022)
\end{botherref}
\endbibitem

\bibitem[\protect\citeauthoryear{Chung et~al.}{2022}]{chung2022scaling}
\begin{botherref}
\oauthor{\bsnm{Chung}, \binits{H.W.}},
\oauthor{\bsnm{Hou}, \binits{L.}},
\oauthor{\bsnm{Longpre}, \binits{S.}},
\oauthor{\bsnm{Zoph}, \binits{B.}},
\oauthor{\bsnm{Tay}, \binits{Y.}},
\oauthor{\bsnm{Fedus}, \binits{W.}},
\oauthor{\bsnm{Li}, \binits{Y.}},
\oauthor{\bsnm{Wang}, \binits{X.}},
\oauthor{\bsnm{Dehghani}, \binits{M.}},
\oauthor{\bsnm{Brahma}, \binits{S.}}, et al.:
Scaling instruction-finetuned language models.
arXiv preprint arXiv:2210.11416
(2022)
\end{botherref}
\endbibitem

\bibitem[\protect\citeauthoryear{Jin et~al.}{2021}]{jin2021disease}
\begin{barticle}
\bauthor{\bsnm{Jin}, \binits{D.}},
\bauthor{\bsnm{Pan}, \binits{E.}},
\bauthor{\bsnm{Oufattole}, \binits{N.}},
\bauthor{\bsnm{Weng}, \binits{W.-H.}},
\bauthor{\bsnm{Fang}, \binits{H.}},
\bauthor{\bsnm{Szolovits}, \binits{P.}}:
\batitle{What disease does this patient have? a large-scale open domain question answering dataset from medical exams}.
\bjtitle{Applied Sciences}
\bvolume{11}(\bissue{14}),
\bfpage{6421}
(\byear{2021})
\end{barticle}
\endbibitem

\bibitem[\protect\citeauthoryear{Manes et~al.}{2024}]{manes2024k}
\begin{botherref}
\oauthor{\bsnm{Manes}, \binits{I.}},
\oauthor{\bsnm{Ronn}, \binits{N.}},
\oauthor{\bsnm{Cohen}, \binits{D.}},
\oauthor{\bsnm{Ber}, \binits{R.I.}},
\oauthor{\bsnm{Horowitz-Kugler}, \binits{Z.}},
\oauthor{\bsnm{Stanovsky}, \binits{G.}}:
K-qa: A real-world medical q\&a benchmark.
arXiv preprint arXiv:2401.14493
(2024)
\end{botherref}
\endbibitem

\bibitem[\protect\citeauthoryear{Toma et~al.}{2023}]{toma2023clinical}
\begin{botherref}
\oauthor{\bsnm{Toma}, \binits{A.}},
\oauthor{\bsnm{Lawler}, \binits{P.R.}},
\oauthor{\bsnm{Ba}, \binits{J.}},
\oauthor{\bsnm{Krishnan}, \binits{R.G.}},
\oauthor{\bsnm{Rubin}, \binits{B.B.}},
\oauthor{\bsnm{Wang}, \binits{B.}}:
Clinical camel: An open-source expert-level medical language model with dialogue-based knowledge encoding.
arXiv preprint arXiv:2305.12031
(2023)
\end{botherref}
\endbibitem

\bibitem[\protect\citeauthoryear{Chen et~al.}{2024}]{chen2024benchmarking}
\begin{botherref}
\oauthor{\bsnm{Chen}, \binits{H.}},
\oauthor{\bsnm{Fang}, \binits{Z.}},
\oauthor{\bsnm{Singla}, \binits{Y.}},
\oauthor{\bsnm{Dredze}, \binits{M.}}:
Benchmarking large language models on answering and explaining challenging medical questions.
arXiv preprint arXiv:2402.18060
(2024)
\end{botherref}
\endbibitem

\bibitem[\protect\citeauthoryear{Wang et~al.}{2023}]{wang2023self}
\begin{bchapter}
\bauthor{\bsnm{Wang}, \binits{X.}},
\bauthor{\bsnm{Wei}, \binits{J.}},
\bauthor{\bsnm{Schuurmans}, \binits{D.}},
\bauthor{\bsnm{Le}, \binits{Q.V.}},
\bauthor{\bsnm{Chi}, \binits{E.H.}},
\bauthor{\bsnm{Narang}, \binits{S.}},
\bauthor{\bsnm{Chowdhery}, \binits{A.}},
\bauthor{\bsnm{Zhou}, \binits{D.}}:
\bctitle{Self-consistency improves chain of thought reasoning in language models}.
In: \bbtitle{The Eleventh International Conference on Learning Representations}
(\byear{2023})
\end{bchapter}
\endbibitem

\bibitem[\protect\citeauthoryear{Nori et~al.}{2023}]{nori2023can}
\begin{botherref}
\oauthor{\bsnm{Nori}, \binits{H.}},
\oauthor{\bsnm{Lee}, \binits{Y.T.}},
\oauthor{\bsnm{Zhang}, \binits{S.}},
\oauthor{\bsnm{Carignan}, \binits{D.}},
\oauthor{\bsnm{Edgar}, \binits{R.}},
\oauthor{\bsnm{Fusi}, \binits{N.}},
\oauthor{\bsnm{King}, \binits{N.}},
\oauthor{\bsnm{Larson}, \binits{J.}},
\oauthor{\bsnm{Li}, \binits{Y.}},
\oauthor{\bsnm{Liu}, \binits{W.}}, et al.:
Can generalist foundation models outcompete special-purpose tuning? case study in medicine.
arXiv preprint arXiv:2311.16452
(2023)
\end{botherref}
\endbibitem

\bibitem[\protect\citeauthoryear{Li et~al.}{2023}]{li2023chatdoctor}
\begin{botherref}
\oauthor{\bsnm{Li}, \binits{Y.}},
\oauthor{\bsnm{Li}, \binits{Z.}},
\oauthor{\bsnm{Zhang}, \binits{K.}},
\oauthor{\bsnm{Dan}, \binits{R.}},
\oauthor{\bsnm{Jiang}, \binits{S.}},
\oauthor{\bsnm{Zhang}, \binits{Y.}}:
Chatdoctor: A medical chat model fine-tuned on a large language model meta-ai (llama) using medical domain knowledge.
Cureus
\textbf{15}(6)
(2023)
\end{botherref}
\endbibitem

\bibitem[\protect\citeauthoryear{Han et~al.}{2023}]{han2023medalpaca}
\begin{botherref}
\oauthor{\bsnm{Han}, \binits{T.}},
\oauthor{\bsnm{Adams}, \binits{L.C.}},
\oauthor{\bsnm{Papaioannou}, \binits{J.-M.}},
\oauthor{\bsnm{Grundmann}, \binits{P.}},
\oauthor{\bsnm{Oberhauser}, \binits{T.}},
\oauthor{\bsnm{L{\"o}ser}, \binits{A.}},
\oauthor{\bsnm{Truhn}, \binits{D.}},
\oauthor{\bsnm{Bressem}, \binits{K.K.}}:
Medalpaca--an open-source collection of medical conversational ai models and training data.
arXiv preprint arXiv:2304.08247
(2023)
\end{botherref}
\endbibitem

\bibitem[\protect\citeauthoryear{Lin}{2004}]{lin2004rouge}
\begin{bchapter}
\bauthor{\bsnm{Lin}, \binits{C.-Y.}}:
\bctitle{Rouge: A package for automatic evaluation of summaries}.
In: \bbtitle{Text Summarization Branches Out},
pp. \bfpage{74}--\blpage{81}
(\byear{2004})
\end{bchapter}
\endbibitem

\bibitem[\protect\citeauthoryear{Zhang et~al.}{2019}]{zhang2019bertscore}
\begin{botherref}
\oauthor{\bsnm{Zhang}, \binits{T.}},
\oauthor{\bsnm{Kishore}, \binits{V.}},
\oauthor{\bsnm{Wu}, \binits{F.}},
\oauthor{\bsnm{Weinberger}, \binits{K.Q.}},
\oauthor{\bsnm{Artzi}, \binits{Y.}}:
Bertscore: Evaluating text generation with bert.
arXiv preprint arXiv:1904.09675
(2019)
\end{botherref}
\endbibitem

\bibitem[\protect\citeauthoryear{Zhang et~al.}{2023}]{zhang2023language}
\begin{botherref}
\oauthor{\bsnm{Zhang}, \binits{M.}},
\oauthor{\bsnm{Press}, \binits{O.}},
\oauthor{\bsnm{Merrill}, \binits{W.}},
\oauthor{\bsnm{Liu}, \binits{A.}},
\oauthor{\bsnm{Smith}, \binits{N.A.}}:
How language model hallucinations can snowball.
arXiv preprint arXiv:2305.13534
(2023)
\end{botherref}
\endbibitem

\bibitem[\protect\citeauthoryear{Ouyang et~al.}{2022}]{ouyang2022training}
\begin{barticle}
\bauthor{\bsnm{Ouyang}, \binits{L.}},
\bauthor{\bsnm{Wu}, \binits{J.}},
\bauthor{\bsnm{Jiang}, \binits{X.}},
\bauthor{\bsnm{Almeida}, \binits{D.}},
\bauthor{\bsnm{Wainwright}, \binits{C.}},
\bauthor{\bsnm{Mishkin}, \binits{P.}},
\bauthor{\bsnm{Zhang}, \binits{C.}},
\bauthor{\bsnm{Agarwal}, \binits{S.}},
\bauthor{\bsnm{Slama}, \binits{K.}},
\bauthor{\bsnm{Ray}, \binits{A.}}, \betal:
\batitle{Training language models to follow instructions with human feedback}.
\bjtitle{Advances in neural information processing systems}
\bvolume{35},
\bfpage{27730}--\blpage{27744}
(\byear{2022})
\end{barticle}
\endbibitem

\bibitem[\protect\citeauthoryear{Pal et~al.}{2022}]{pal2022medmcqa}
\begin{bchapter}
\bauthor{\bsnm{Pal}, \binits{A.}},
\bauthor{\bsnm{Umapathi}, \binits{L.K.}},
\bauthor{\bsnm{Sankarasubbu}, \binits{M.}}:
\bctitle{Medmcqa: A large-scale multi-subject multi-choice dataset for medical domain question answering}.
In: \bbtitle{Conference on Health, Inference, and Learning},
pp. \bfpage{248}--\blpage{260}
(\byear{2022}).
\bcomment{PMLR}
\end{bchapter}
\endbibitem

\bibitem[\protect\citeauthoryear{Abacha et~al.}{2017}]{abacha2017overview}
\begin{bchapter}
\bauthor{\bsnm{Abacha}, \binits{A.B.}},
\bauthor{\bsnm{Agichtein}, \binits{E.}},
\bauthor{\bsnm{Pinter}, \binits{Y.}},
\bauthor{\bsnm{Demner-Fushman}, \binits{D.}}:
\bctitle{Overview of the medical question answering task at trec 2017 liveqa.}
In: \bbtitle{TREC},
pp. \bfpage{1}--\blpage{12}
(\byear{2017})
\end{bchapter}
\endbibitem

\bibitem[\protect\citeauthoryear{Abacha et~al.}{2019}]{abacha2019bridging}
\begin{bchapter}
\bauthor{\bsnm{Abacha}, \binits{A.B.}},
\bauthor{\bsnm{Mrabet}, \binits{Y.}},
\bauthor{\bsnm{Sharp}, \binits{M.}},
\bauthor{\bsnm{Goodwin}, \binits{T.R.}},
\bauthor{\bsnm{Shooshan}, \binits{S.E.}},
\bauthor{\bsnm{Demner-Fushman}, \binits{D.}}:
\bctitle{Bridging the gap between consumers' medication questions and trusted answers.}
In: \bbtitle{MedInfo},
pp. \bfpage{25}--\blpage{29}
(\byear{2019})
\end{bchapter}
\endbibitem

\bibitem[\protect\citeauthoryear{Abacha et~al.}{2023}]{abacha2023empirical}
\begin{bchapter}
\bauthor{\bsnm{Abacha}, \binits{A.B.}},
\bauthor{\bsnm{Yim}, \binits{W.-w.}},
\bauthor{\bsnm{Fan}, \binits{Y.}},
\bauthor{\bsnm{Lin}, \binits{T.}}:
\bctitle{An empirical study of clinical note generation from doctor-patient encounters}.
In: \bbtitle{Proceedings of the 17th Conference of the European Chapter of the Association for Computational Linguistics},
pp. \bfpage{2283}--\blpage{2294}
(\byear{2023})
\end{bchapter}
\endbibitem

\bibitem[\protect\citeauthoryear{Zhang et~al.}{2023}]{zhang2023alpacare}
\begin{botherref}
\oauthor{\bsnm{Zhang}, \binits{X.}},
\oauthor{\bsnm{Tian}, \binits{C.}},
\oauthor{\bsnm{Yang}, \binits{X.}},
\oauthor{\bsnm{Chen}, \binits{L.}},
\oauthor{\bsnm{Li}, \binits{Z.}},
\oauthor{\bsnm{Petzold}, \binits{L.R.}}:
Alpacare: Instruction-tuned large language models for medical application.
arXiv preprint arXiv:2310.14558
(2023)
\end{botherref}
\endbibitem

\bibitem[\protect\citeauthoryear{Alberti et~al.}{2019}]{alberti-etal-2019-synthetic}
\begin{bchapter}
\bauthor{\bsnm{Alberti}, \binits{C.}},
\bauthor{\bsnm{Andor}, \binits{D.}},
\bauthor{\bsnm{Pitler}, \binits{E.}},
\bauthor{\bsnm{Devlin}, \binits{J.}},
\bauthor{\bsnm{Collins}, \binits{M.}}:
\bctitle{Synthetic {QA} corpora generation with roundtrip consistency}.
In: \bbtitle{Proceedings of the 57th Annual Meeting of the Association for Computational Linguistics},
pp. \bfpage{6168}--\blpage{6173}.
\bpublisher{Association for Computational Linguistics},
\blocation{Florence, Italy}
(\byear{2019}).
\doiurl{10.18653/v1/P19-1620} .
\burl{https://aclanthology.org/P19-1620}
\end{bchapter}
\endbibitem

\bibitem[\protect\citeauthoryear{Sattar}{2020}]{sattar2020fundamentals}
\begin{bbook}
\bauthor{\bsnm{Sattar}, \binits{H.A.}}:
\bbtitle{Fundamentals of Pathology: Medical Course and Step 1 Review}.
\bpublisher{Pathoma.com}, \blocation{???}
(\byear{2020})
\end{bbook}
\endbibitem

\bibitem[\protect\citeauthoryear{Wang et~al.}{2022}]{wang2022self}
\begin{botherref}
\oauthor{\bsnm{Wang}, \binits{Y.}},
\oauthor{\bsnm{Kordi}, \binits{Y.}},
\oauthor{\bsnm{Mishra}, \binits{S.}},
\oauthor{\bsnm{Liu}, \binits{A.}},
\oauthor{\bsnm{Smith}, \binits{N.A.}},
\oauthor{\bsnm{Khashabi}, \binits{D.}},
\oauthor{\bsnm{Hajishirzi}, \binits{H.}}:
Self-instruct: Aligning language model with self generated instructions.
arXiv preprint arXiv:2212.10560
(2022)
\end{botherref}
\endbibitem

\bibitem[\protect\citeauthoryear{Taori et~al.}{2023}]{alpaca}
\begin{botherref}
\oauthor{\bsnm{Taori}, \binits{R.}},
\oauthor{\bsnm{Gulrajani}, \binits{I.}},
\oauthor{\bsnm{Zhang}, \binits{T.}},
\oauthor{\bsnm{Dubois}, \binits{Y.}},
\oauthor{\bsnm{Li}, \binits{X.}},
\oauthor{\bsnm{Guestrin}, \binits{C.}},
\oauthor{\bsnm{Liang}, \binits{P.}},
\oauthor{\bsnm{Hashimoto}, \binits{T.B.}}:
Stanford Alpaca: An Instruction-following LLaMA model.
GitHub
(2023)
\end{botherref}
\endbibitem

\bibitem[\protect\citeauthoryear{Dao et~al.}{2022}]{dao2022flashattention}
\begin{barticle}
\bauthor{\bsnm{Dao}, \binits{T.}},
\bauthor{\bsnm{Fu}, \binits{D.}},
\bauthor{\bsnm{Ermon}, \binits{S.}},
\bauthor{\bsnm{Rudra}, \binits{A.}},
\bauthor{\bsnm{R{\'e}}, \binits{C.}}:
\batitle{Flashattention: Fast and memory-efficient exact attention with io-awareness}.
\bjtitle{Advances in Neural Information Processing Systems}
\bvolume{35},
\bfpage{16344}--\blpage{16359}
(\byear{2022})
\end{barticle}
\endbibitem

\bibitem[\protect\citeauthoryear{Hendrycks et~al.}{2020}]{hendrycks2020measuring}
\begin{botherref}
\oauthor{\bsnm{Hendrycks}, \binits{D.}},
\oauthor{\bsnm{Burns}, \binits{C.}},
\oauthor{\bsnm{Basart}, \binits{S.}},
\oauthor{\bsnm{Zou}, \binits{A.}},
\oauthor{\bsnm{Mazeika}, \binits{M.}},
\oauthor{\bsnm{Song}, \binits{D.}},
\oauthor{\bsnm{Steinhardt}, \binits{J.}}:
Measuring massive multitask language understanding.
arXiv preprint arXiv:2009.03300
(2020)
\end{botherref}
\endbibitem

\bibitem[\protect\citeauthoryear{Min et~al.}{2023}]{min2023factscore}
\begin{botherref}
\oauthor{\bsnm{Min}, \binits{S.}},
\oauthor{\bsnm{Krishna}, \binits{K.}},
\oauthor{\bsnm{Lyu}, \binits{X.}},
\oauthor{\bsnm{Lewis}, \binits{M.}},
\oauthor{\bsnm{Yih}, \binits{W.-t.}},
\oauthor{\bsnm{Koh}, \binits{P.W.}},
\oauthor{\bsnm{Iyyer}, \binits{M.}},
\oauthor{\bsnm{Zettlemoyer}, \binits{L.}},
\oauthor{\bsnm{Hajishirzi}, \binits{H.}}:
Factscore: Fine-grained atomic evaluation of factual precision in long form text generation.
arXiv preprint arXiv:2305.14251
(2023)
\end{botherref}
\endbibitem

\bibitem[\protect\citeauthoryear{Kwon et~al.}{2023}]{kwon2023efficient}
\begin{bchapter}
\bauthor{\bsnm{Kwon}, \binits{W.}},
\bauthor{\bsnm{Li}, \binits{Z.}},
\bauthor{\bsnm{Zhuang}, \binits{S.}},
\bauthor{\bsnm{Sheng}, \binits{Y.}},
\bauthor{\bsnm{Zheng}, \binits{L.}},
\bauthor{\bsnm{Yu}, \binits{C.H.}},
\bauthor{\bsnm{Gonzalez}, \binits{J.E.}},
\bauthor{\bsnm{Zhang}, \binits{H.}},
\bauthor{\bsnm{Stoica}, \binits{I.}}:
\bctitle{Efficient memory management for large language model serving with pagedattention}.
In: \bbtitle{Proceedings of the ACM SIGOPS 29th Symposium on Operating Systems Principles}
(\byear{2023})
\end{bchapter}
\endbibitem

\bibitem[\protect\citeauthoryear{Ko et~al.}{2020}]{ko2020look}
\begin{botherref}
\oauthor{\bsnm{Ko}, \binits{M.}},
\oauthor{\bsnm{Lee}, \binits{J.}},
\oauthor{\bsnm{Kim}, \binits{H.}},
\oauthor{\bsnm{Kim}, \binits{G.}},
\oauthor{\bsnm{Kang}, \binits{J.}}:
Look at the first sentence: Position bias in question answering.
arXiv preprint arXiv:2004.14602
(2020)
\end{botherref}
\endbibitem

\end{thebibliography}
